\documentclass[acmsmall, nonacm]{acmart}

\usepackage[utf8]{inputenc}
\usepackage[american]{babel}
\usepackage{graphicx}
\usepackage{amsmath}
\usepackage{mathtools}
\usepackage{amsfonts}
\usepackage{bm}
\usepackage{array}
\usepackage{subcaption}
\usepackage[]{algorithm2e}
\usepackage[inline,shortlabels]{enumitem}
\usepackage{lscape}
\usepackage{multirow}
\usepackage{longtable}
\usepackage{newtxmath}
\usepackage{tikz}
\usetikzlibrary{trees,arrows, chains, fit, positioning, calc, shapes, shadows, arrows.meta, intersections, shadows.blur, patterns, patterns.meta}
\tikzset{narrowtriangle/.tip={Triangle[length = 2pt 1.5, width=4pt 2, round, line width = 1pt 1]}}

\DeclareMathOperator*{\argmax}{argmax}
\DeclareMathOperator{\EX}{\mathbb{E}}

\ifdefined\N                                                                
\renewcommand{\N}{\vmathbb{N}}                                                
\else
  \newcommand{\N}{\vmathbb{N}}
\fi
\newcommand{\R}{\vmathbb{R}}                                                 
\ifdefined\C
  \renewcommand{\C}{\vmathbb{C}}                                             
\else
  \newcommand{\C}{\vmathbb{C}}
\fi

\def\argmax{\mathop{\sf arg\,max}}                                          
\def\argmin{\mathop{\sf arg\,min}}                                          

\newcommand{\xv}{\mathbf{x}}													
\newcommand{\xb}{\mathbf{x}}													

\renewcommand{\P}{\vmathbb{P}}                                               
\newcommand{\E}{\vmathbb{E}}                                                 

\newcommand{\Xspace}{\mathcal{X}}                                           
\newcommand{\Yspace}{\mathcal{Y}}                                           
\newcommand{\Pxy}{\P_{xy}}                                                  
\newcommand{\xy}{(\mathbf{x}, y)}                                                  
\newcommand{\D}{\mathcal{D}}                                                
\renewcommand{\xi}[1][i]{\mathbf{x}^{(#1)}}                                          
\newcommand{\yi}[1][i]{y^{(#1)}}                                            
\newcommand{\Dtrain}{\mathcal{D}_{\text{train}}}                            
\newcommand{\Dtest}{\mathcal{D}_{\text{test}}}                              

\newcommand{\inducer}{\mathcal{I}}                                                

\newcommand{\fx}{f(\mathbf{x})}                                                      
\newcommand{\Hspace}{\mathcal{H}}														
\newcommand{\Lxy}{L\left(y, \fx\right)}                                               

\newcommand{\riske}{\mathcal{R}_{\text{emp}}}                               
\newcommand{\riskef}{\riske(f)}                                             

\newcommand{\lambdav}{\bm{\lambda}}											

\newcommand{\lambdab}{\boldsymbol{\lambda}}									

\newcommand{\LamS}{\tilde\Lambda}                           
\newcommand{\archive}{\mathcal{A}}                          

\newcommand{\chlam}{\hat{c}(\lambdav)}                       
\newcommand{\shlam}{\hat{\sigma}(\lambdav)}                  
\newcommand{\ulam}{u(\lambdav)}                              

\AtBeginDocument{%
  \providecommand\BibTeX{{%
    \normalfont B\kern-0.5em{\scshape i\kern-0.25em b}\kern-0.8em\TeX}}}

\makeatletter
\renewcommand*{\citet}[2][, ]{%
  \def\nextitem{\def\nextitem{#1}}%
  \@for \el:=#2\do{\nextitem\citeauthor{\el}~[\citeyear{\el}]}%
}
\makeatother

\copyrightyear{2022}
\acmYear{2022}
\acmDOI{XXXXXXX.XXXXXXX}

\acmConference[Conference acronym 'XX]{Make sure to enter the correct
  conference title from your rights confirmation emai}{June 03--05,
  2018}{Woodstock, NY}
\acmPrice{15.00}
\acmISBN{978-1-4503-XXXX-X/18/06}
\acmJournal{TELO}
\acmVolume{TBD}
\acmNumber{TBD}
\acmMonth{0}

\begin{document}

\title{Multi-Objective Hyperparameter Optimization in Machine Learning -- An Overview}


\author{Florian Karl}
\authornote{Both authors contributed equally to this research.}
\email{florian.karl@iis.fraunhofer.de}
\affiliation{%
  \institution{Fraunhofer Institut für integrierte Schaltungen}
  \country{Germany}
}

\author{Tobias Pielok}
\authornotemark[1]
\email{tobias.pielok@stat.uni-muenchen.de}
\affiliation{%
  \institution{Ludwig-Maximilians-Universität München}
  \country{Germany}
}

\author{Julia Moosbauer}
\affiliation{%
  \institution{Ludwig-Maximilians-Universität München}
  \country{Germany}
}

\author{Florian Pfisterer}
\affiliation{%
  \institution{Ludwig-Maximilians-Universität München}
  \country{Germany}
}

\author{Stefan Coors}
\affiliation{%
  \institution{Ludwig-Maximilians-Universität München}
  \country{Germany}
}

\author{Martin Binder}
\affiliation{%
  \institution{Ludwig-Maximilians-Universität München}
  \country{Germany}
}

\author{Lennart Schneider}
\affiliation{%
  \institution{Ludwig-Maximilians-Universität München}
  \country{Germany}
}

\author{Janek Thomas}
\affiliation{%
  \institution{Ludwig-Maximilians-Universität München}
  \country{Germany}
}

\author{Jakob Richter}
\affiliation{%
  \institution{Ludwig-Maximilians-Universität München}
  \country{Germany}
}

\author{Michel Lang}
\affiliation{%
  \institution{Technische Universität Dortmund}
  \country{Germany}
}

\author{Eduardo C. Garrido-Merch\'an}
\affiliation{%
  \institution{Universidad Pontificia Comillas}
  \country{Spain}
}

\author{Juergen Branke}
\affiliation{%
  \institution{Warwick Business School}
  \country{UK}
}

\author{Bernd Bischl}
\affiliation{%
  \institution{Ludwig-Maximilians-Universität München}
  \country{Germany}
}


\renewcommand{\shortauthors}{Karl and Pielok, et al.}

\begin{abstract}
Hyperparameter optimization constitutes a large part of typical modern machine learning workflows. This arises from the fact that machine learning methods and corresponding preprocessing steps often only yield optimal performance when hyperparameters are properly tuned.
But in many applications, we are not only interested in optimizing ML pipelines solely for predictive accuracy; additional metrics or constraints must be considered when determining an optimal configuration, resulting in a multi-objective optimization problem. 
This is often neglected in practice, due to a lack of knowledge and readily available software implementations for multi-objective hyperparameter optimization. 
In this work, we introduce the reader to the basics of multi-objective hyperparameter optimization and motivate its usefulness in applied ML.
Furthermore, we provide an extensive survey of existing optimization strategies, both from the domain of \textit{evolutionary algorithms} and \textit{Bayesian optimization}. 
We illustrate the utility of MOO in several specific ML applications, considering objectives such as operating conditions, prediction time, sparseness, fairness, interpretability and robustness.
\end{abstract}



\maketitle

\section{Introduction}

With the immense popularity of machine learning (ML) and data-driven solutions for many domains~\citep{sarker2021machine}, the demand for automating the creation of suitable ML pipelines has strongly increased~\citep{automl_book}. 
Automated machine learning (AutoML) and hyperparameter optimization (HPO) promise to simplify the ML process by enabling less experienced practitioners to optimally configure ML models for a variety of tasks - reducing manual effort and improving performance at the same time~\citep{bischl2021hyperparameter, he2020automl, yao2018taking, luo2016review, automl_book}.
HPO is often classified as a black-box optimization problem, a type of optimization problem where only the outputs of the function to be optimized can be observed and no analytic expression of the underlying function is known. 
Furthermore, these optimization problems are often noisy and expensive~\citep{audet2017derivative}.
Another challenge in HPO is that there is rarely a clear-cut, obvious, single  performance metric, which is in stark contrast to the setup often described in many algorithm-centric research papers and introductory books~\citep{he2020automl, lu2020nsganetv2, automl_book}. 

Nowadays, models and pipelines are held to a high standard,
as the ML process (as currently realized in many businesses and institutions) comes with a number of different stakeholders.
While predictive performance measures are still decisive in most cases,
models must be reliable, robust, accountable for their decisions, efficient for seamless deployment, and so on.
In many internet-of-things applications, an ML model is deployed on edge devices like smartphones, watches, and embedded systems~\citep{howard2017mobilenets}. 
Power consumption, memory capacity, and inference latency can be limiting factors when deploying models in such settings, and obvious trade-offs between these factors and predictive performance exist.
Even 
expressing just predictive performance as a single metric can frequently be challenging. 
The best known example of this is perhaps in medical applications:
For a diagnostic test, solely looking at misclassification rates is ill-advised: Misclassifying a sick patient as healthy (false negative) has usually much more severe consequences than classifying a healthy person erroneously as sick (false positive), i.e., different \emph{misclassification costs}, which are often unknown or hard to quantify, have to be considered~\citep{fawcett2006introduction}.

Of course, multiple objectives can in principle be aggregated into a single metric, which converts a multi-objective optimization (MOO) problem to a single-objective optimization problem.
However, it is often unclear how a trade-off between different objectives should be defined \textit{a priori}, i.e., before possible alternative solutions are known
~\citep{brodley2012challenges}.
In this paper, we argue that there is substantial merit in directly approaching a multi-objective HPO (MOHPO) problem as such and give a comprehensive review of methods, tools and applications.
Furthermore, other prominent work in ML research - aside from HPO - has advocated for a multi-objective perspective~\citep{jin2006multi, iqbal2020flexibo, mierswa_thesis2019}.
For example, as argued in~\citet{mierswa_thesis2019}, many ML and data mining applications inherently concern trade-offs and thus should be approached via MOO methods. And even if the main interest lies in a single objective it still might be advantageous to approach the problem via MOO methods since they have the potential to reduce local minima ~\citep{knowles2001reducingloc}. 
MOO algorithms seek to approximate the set of \emph{efficient} or \emph{Pareto-optimal} solutions. These solutions have different trade-offs, but it is not possible to improve any objective without degrading at least one other objective. 
This set of Pareto optimal solutions can then be analyzed by domain experts in a post-hoc manner, and an informed decision can be made as to which trade-off should be used in the application, without requiring the user to specify this a priori~\citep{jin2008pareto, jin2006multi}.\\
\\
This paper provides a comprehensive review on the topic of MOHPO, explains the most popular algorithms, discusses main challenges and opportunities, and surveys existing applications. 
We restrict the scope of this paper to the realm of supervised ML.
Unsupervised ML, in contrast, entails a different set of metrics to the scenario studied in our manuscript and is largely governed by custom, use case-specific measures~\citep{palacio2019evaluation, fan2020hyperparameter} and sometimes even visual inspection of results.
We address this review mainly towards ML practitioners looking for a comprehensive introduction and review of MOO for HPO; though this work may also be of value to researchers and practitioners familiar with MOO and interested in applying their skill set in the field of ML.
The rest of the paper is structured as follows. 
First we define the MOHPO problem in Section~\ref{sec:problem_defintion}. 
Section~\ref{sec:foundations} presents the theoretical foundations of MOO, i.e., how to evaluate sets of candidate solutions. 
Then, we introduce several important MOHPO methods in Section~\ref{sec:multi_objective_methods}, such as multi-objective evolutionary algorithms (MOEAs), Bayesian Optimization (BO) and also Hyperband.
Finally, Section~\ref{sec:applications} introduces a number of applications for MOHPO and their associated objectives.
We will categorize these applications through exploring three overarching perspectives on the ML process: (1) Performance metrics, (2) metrics that measure costs and restrictions at deployment like efficiency, and (3) metrics that enforce reliability and interpretability.

\section{Hyperparameter optimization}
\label{sec:problem_defintion}

\subsection{The machine learning problem}
\tikzset{
  basic/.style  = {draw, text width=4cm, drop shadow, font=\sffamily \scriptsize, rectangle},
  root/.style   = {basic, rounded corners=2pt, thin, align=center, fill=white},
  level-2/.style = {basic, rounded corners=5pt, thin,align=center, fill=white, text width=2.3cm},
  level-3/.style = {basic, thin, align=center, fill=white, text width=1.8cm},
  level-3-focus/.style = {basic, thick, align=center, fill=white, text width=1.8cm},
  level-4/.style = {basic, thin, align=center, fill=white, text width=1.6cm},
  level-4-focus/.style = {basic, thick, align=center, fill=white, text width=1.6cm}
}

\begin{figure}[ht]
  \centering
  \begin{tikzpicture}[scale=0.55, every node/.style={transform shape},
    shorten >=0.5pt, >={Stealth[round]},
    hbox/.style = {rectangle, rounded corners = 2, node distance = 2.5cm, fill = white, draw = black, thick, text width = 10em, align = center,
        blur shadow = {shadow blur steps = 5}, minimum height = 1cm, minimum width = 1.5cm, font = \large
},
    sbox/.style = {rectangle, rounded corners = 2, node distance = 1cm, fill = white, draw = black, thick, text width = 8em, align = center,
        blur shadow = {shadow blur steps = 5}, minimum height = 1cm, minimum width = 1.5cm
}]

\node[hbox, text width = 20em] (a) at (0, 0) {\textbf{Taxonomy of black-box problems}};
\node[hbox, below of = a, xshift = -7.5cm] (b) {\textbf{Stochasticity}};
\node[hbox, below of = a, xshift = -2.5cm] (c) {\textbf{Domain}};
\node[hbox, below of = a, xshift = 2.5cm] (d) {\textbf{Codomain}};
\node[hbox, below of = a, xshift = 7.5cm] (e) {\textbf{Evaluation cost}};
\draw[-narrowtriangle, line width = 0.3mm, black] (a.south) -- (0, -1) -- (b.north |- , -1)  -- (b.north);
\draw[-narrowtriangle, line width = 0.3mm, black] (a.south) -- (0, -1) -- (c.north |- , -1)  -- (c.north);
\draw[-narrowtriangle, line width = 0.3mm, black] (a.south) -- (0, -1) -- (d.north |- , -1)  -- (d.north);
\draw[-narrowtriangle, line width = 0.3mm, black] (a.south) -- (0, -1) -- (e.north |- , -1)  -- (e.north);

\node[sbox, below right of = b, yshift = -1.5cm] (b1) {deterministic};
\node[sbox, below right of = b, yshift = -3.0cm] (b2) {stochastic, homoscedastic};
\node[sbox, fill = green!60!black, below right of = b, yshift = -4.5cm] (b3) {\textbf{stochastic, heteroscedastic}};
\foreach \value in {1, 2, 3}
    \draw[-narrowtriangle, line width = 0.3mm, black] (b.text |- b.south) |- (b\value.west);

\node[sbox, below right of = c, yshift = -1.5cm] (c1) {dimensionality};
\node[sbox, below right of = c, yshift = -7cm] (c2) {type};
\foreach \value in {1, 2}
    \draw[-narrowtriangle, line width = 0.3mm, black] (c.text |- c.south) |- (c\value.west);
    
\node[sbox, below right of = c1, yshift = -1.5cm] (c11) {low dim $d$};
\node[sbox, below right of = c1, yshift = -3cm] (c12) {high dim $d$};
\foreach \value in {1, 2}
    \draw[-narrowtriangle, line width = 0.3mm, black] (c1.text |- c1.south) |- (c1\value.west);

\node[sbox, below right of = c2, yshift = -1.5cm] (c21) {numerical};
\node[sbox, below right of = c2, yshift = -3.0cm] (c22) {categorical};
\node[sbox, fill = green!60!black, below right of = c2, yshift = -4.5cm] (c23) {\textbf{mixed numerical and categorical}};
\node[sbox, fill = green!60!black, below right of = c2, yshift = -6.0cm] (c24) {\textbf{hierarchical /\\structured}};
\foreach \value in {1, 2, 3, 4}
    \draw[-narrowtriangle, line width = 0.3mm, black] (c2.text |- c2.south) |- (c2\value.west);

\node[sbox, below right of = d, yshift = -1.5cm] (d1) {single-objective $\R$};
\node[sbox, fill = green!60!black, below right of = d, yshift = -3.0cm] (d2) {\textbf{multi-objective $\R^m, m \in \{2,3,4\}$}};
\node[sbox, below right of = d, yshift = -4.5cm] (d3) {many objective $\R^{m}, m \ge 5$};
\foreach \value in {1, 2, 3}
    \draw[-narrowtriangle, line width = 0.3mm, black] (d.text |- d.south) |- (d\value.west);

\node[sbox, below right of = e, yshift = -1.5cm] (e1) {cheap};
\node[sbox, fill = green!60!black, below right of = e, yshift = -3cm] (e2) {\textbf{expensive}};
\foreach \value in {1, 2}
    \draw[-narrowtriangle, line width = 0.3mm, black] (e.text |- e.south) |- (e\value.west);

\end{tikzpicture}
  \caption{Taxonomy of common black-box optimization problems. Attributes that are related to MOHPO - and therefore a substantial focus in this paper - are highlighted in green.}
  \Description{Division of black-box problems into stochasticity, domain, codomain and evaluation cost. Division of stochasticity into deterministic, stochastic (homescedastic) and stocastic (heteroscedastic). Division of domain into dimensionality (divided into low dim and high dim) and type (divided into numerical, purely categorical, mixed numerical and categorical, hierarchical/structured). Division of codomain into single-objective, multi-objective and many objective (more than 4 objectives). Division of evaluation cost into cheap and expensive.}
  \label{fig:taxonomy_problem_classes}
\end{figure}
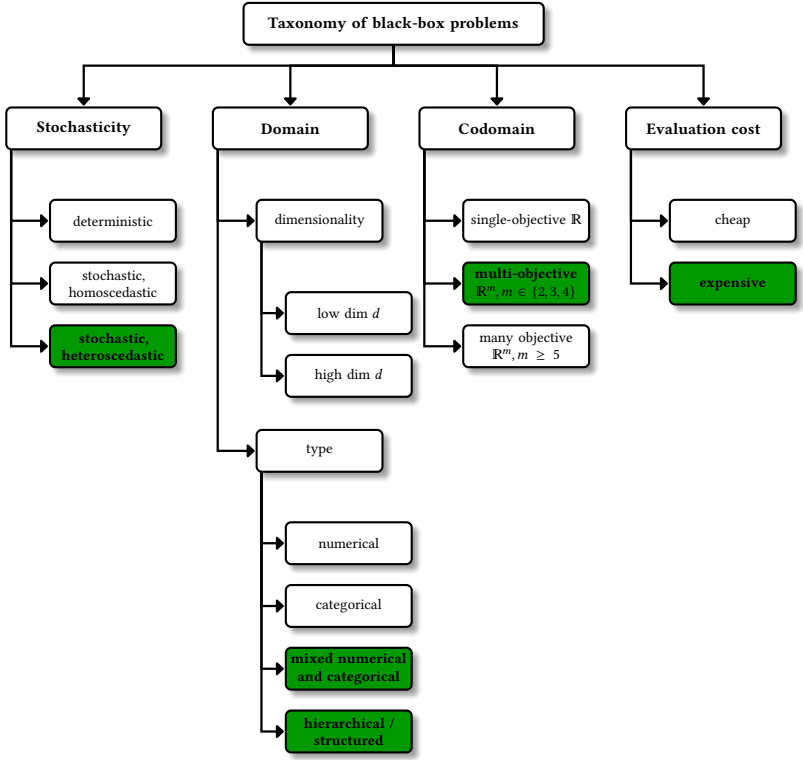

The fundamental ML problem can be defined as follows.
Let $\D$ be a dataset with $n$ input-output pairs $\left(\xi, \yi\right) \in \Xspace \times \Yspace$, which are independent and identically distributed (i.i.d.) from a data-generating distribution $\Pxy$. 
An ML algorithm $\inducer\left(\cdot, \lambdab\right)$ configured by hyperparameters $\lambdab \in \Lambda$ maps a dataset $\D$ to a model $f: \Xspace \to \R^g$ where $\R^g$ with $g \in \mathbb{N}$ is the numerical encoded space\footnote{E.g., for regression, $g = 1$. For classification, $g$ is the number of classes and given a certain input, $\inducer$ outputs a probability score for each class .} of $\Yspace$  via
$
\inducer: \left(\vmathbb{D} \times \Lambda\right) \to \Hspace, 
$
where $\Hspace$ is the hypothesis space which consists of all models and $\vmathbb{D}$ 
is the set of all finite datasets. With a slight abuse of notation we will also write $\inducer_{\lambdav}$ if the hyperparameter $\lambdav$ is fixed, i.e., $\inducer_{\lambdav}(\D) = \inducer(\D, \lambdav).$
What we would like to optimize is the expected generalization performance $GE$ of our model w.r.t. a loss $L$, when trained on $\D$, on new unseen data $\xy \notin \D$, i.e.,
\begin{eqnarray}
\mathrm{GE}(\inducer, \lambdav, n, L) = \E_{(\xv, y) \sim \Pxy, \D \sim \Pxy^n} \left[L(y, \inducer_{\lambdav}(\D)(\xv)) \right]
\end{eqnarray}
with the expectation taken over the random data $\D$ of size $n$ and a fresh test sample $(\xv,y)$, both independently sampled from $\Pxy$. 
Since the data generating distribution $\Pxy$ is usually unknown, also GE has to be estimated. 
To do so, the data $\D$ is split into $B$ training and test sets by a resampling method $\Dtrain^b$ and $ \Dtest^b$ such that $\D = \Dtrain^b \cup~ \Dtest^b, b = 1, \dots, B$.
The expected generalization error can then be computed as 
\begin{eqnarray}
\label{eq:GEH}
        \widehat{\mathrm{GE}}\left(\inducer, \mathcal{J}, L, \lambdab\right) := \frac{1}{B} \sum_{b = 1}^{B} \frac{1}{|\Dtest^b|} \sum_{\xy \in \Dtest^b} L\left(y, \inducer_{\lambdav}(\Dtrain^b)(\xv)\right) 
\end{eqnarray}
where $\mathcal{J}$ is the set of train-test-splits.
Often, learners $\inducer$ internally minimize a regularized error w.r.t. a point-wise loss function $\Lxy: \Yspace\times\R^g \to \R$, i.e.,
\begin{eqnarray}\label{eq:ml_riskmin}
    \inducer_{\lambdab}(\D) = \argmin_{f \in \Hspace} \riskef +  \eta  J(f), \quad \textrm{ with } \quad \riskef := \sum_{(\xb, y) \in \D} \Lxy
\end{eqnarray} 
where $\riske$ is the so-called empirical risk.
The regularizer $J(f)$ expresses a preference (or prior in the Bayesian sense) for simpler models and is usually formulated as some kind of norm on the parameter vector, e.g., the L2 norm for a ridge penalty, and its strength is controlled by a hyperparameter $\eta$
\footnote{This can already be seen as a simple scalarization of a multi-objective problem, as we combine the different measures for empirical risk and regularization into a weighted sum.~\citet{mierswa_thesis2019} derives the two conflicting objectives from this scalarized version and presents an explicit multi-objective formulation of the optimization problem (minimizing the regularized empirical risk).}.
Most learning algorithms are configured by a possibly large number of hyperparameters $\lambdab$ that control the hypothesis space or the fitting procedure. 
Usually, the generalization error $\widehat{\mathrm{GE}}\left(\inducer, \mathcal{J}, L, \lambdab\right)$ critically depends on the choice of $\lambdab$. 
In most cases, the analytical relationship of hyperparameters and generalization error is unknown, and even to experts it is not clear how to choose optimal hyperparameter values. 
Hyperparameter optimization aims to minimize the estimated generalization error $\widehat{\mathrm{GE}}$ for a given dataset $\D$ using the hyperparameter configuration (HPC) $\lambdab$ =
$
     \argmin_{\lambdab \in \LamS} \widehat{\mathrm{GE}}\left(\inducer,  \lambdab\right), 
$     
where $\LamS$, a bounded subspace of the hyperparameter space $\Lambda$, is the so-called \emph{search-space} or \emph{domain} in the context of the above optimization problem. 
Note that it is important to not use any test data during training or HPO as this could lead to an optimistic bias.
Instead, nested resampling techniques should be applied~\citep{bischl2012resampling}: 
The data is split into an optimization set $\mathcal{D}_{\textrm{optim}}$ and a test set $\mathcal{D}_{\textrm{test}}$. 
During the optimization process, the model performance should only be assessed by an (inner) resampling technique, e.g., cross-validation, on $\mathcal{D}_{\textrm{optim}}$, whereas $\mathcal{D}_{\textrm{test}}$ should only be used for the final assessment of the chosen model(s).
A simple alternative is simply splitting $\mathcal{D}_{\textrm{optim}}$ into two datasets $\mathcal{D}_{\textrm{train}}$ and $\mathcal{D}_{\textrm{val}}$, which leads to the widely known train/validation/test-split.
Afterwards, a resulting non-dominated solution set can be determined on $\mathcal{D}_{\textrm{test}}$ based on the solution candidates found on $\mathcal{D}_{\textrm{optim}}$. 
In practice, it is very important to have no hidden information leakage from test set to training, which may for example happen when preprocessing (such as missing value imputation or feature selection) is done on the combined set.
The estimate based on a single train-test-split can have a high variance, for example because of a strong dependence on the data split. 
To counter this, an (outer) resampling strategy like cross-validation is usually used, and the average of the estimated generalizations of the quality indicator on 
is reported.

\subsection{Multi-objective hyperparemter optimization as a black box optimization problem}

As there is generally no analytical expression of the general hyperparameter optimization problem, it forms a black-box function. Black-box problems can be characterized according to different attributes (see Figure~\ref{fig:taxonomy_problem_classes}), which influence the difficulty of the problem. 
For further fundamental details on single-objective optimization for HPO see~\citep{bischl2021hyperparameter}.
In an ML scenario, we are often interested in the generalization error with respect to more than one loss as well as further relevant criteria like robustness, interpretability, sparseness, or efficiency of a resulting model~\citep{jin2006multi}. 
Therefore, we introduce a generalized definition of the hyperparameter optimization problem that takes into account $m$ criteria: 
Given a number of evaluation criteria $c_1: \tilde{\Lambda} \to \R, \dots, c_m: \tilde{\Lambda} \to \R$ with $m \in \N$, we define $c: \tilde{\Lambda} \to \R^m$ to assign an $m$-dimensional cost vector to a HPC $\lambdab$. 
The estimated generalization error $\widehat{\mathrm{GE}}$ is one evaluation criterion that is commonly used, but many further criteria will be discussed in this paper.
The general multi-objective hyperparameter optimization problem can be defined as 
\begin{align}\label{eq:multi_obj}
  \argmin_{\lambdab \in \tilde{\Lambda}} c(\lambdab) = \argmin_{\lambdab \in \tilde{\Lambda}} \left( c_1(\lambdab), c_2(\lambdab), \ldots, c_m(\lambdab) \right) \ .
\end{align}

Without loss of generality, we assume that all criteria are minimized. 
The domain $\Lambda$ of the problem is called numerical if only numeric hyperparameters $\lambdab$ are optimized. 
By including additional categorical hyperparameters, like the type of kernel used in a support sector machine (SVM), the search space becomes \textit{mixed} numerical and categorical.
Mixed search spaces already require adaption of some optimization strategies, such as BO, which we will discuss in Section~\ref{ssec:model_based_optimization}.
It can also be necessary to introduce further conditional hierarchies between hyperparameters.
For example, when optimizing over different kernel types of an SVM, the $\gamma$ kernel hyperparameter is only valid if the kernel type is set to \emph{Radial Basis Function} (RBF), while for a polynomial kernel, a hyperparameter for the polynomial degree must be specified. 
These conditional hierarchies can become highly complicated - especially when moving from pure HPO to optimizing over full ML pipelines, i.e., AutoML, or over neural network architectures, referred to as neural architecture search (NAS)~\citep{pham2018efficient, feurer2015efficient, olson2016tpot}.
The dimensionality of the search space $\dim(\LamS)$ also directly influences the difficulty of the problem: 
While it might be desirable to include as many hyperparameters as possible (since it is often unknown which hyperparameters are actually important), this increases the complexity of the optimization problem and requires an increasingly larger budget of (expensive) evaluations. 
The \textit{co-domain}, also called objective space, i.e., $\R^m$, of the problem is characterized by the number of objectives $m$. 
Objective functions $c_i, i \in \{1, 2, ..., m\}$ can be characterized by their \textit{evaluation cost} and \textit{stochasticity}.
Some evaluation criteria are deterministic, such as the required memory of a model on hard disk.
Other criteria can only be measured with some additional noise, e.g., the generalization performance discussed above.
To summarize, MOHPO is an expensive, multi-objective optimization problem on a mixed and hierarchical search space that is possibly heteroscedastic in its objectives; the domain can be either low dimensional or high dimensional. 
A visualization of this can be found in Figure~\ref{fig:taxonomy_problem_classes}.
We will give a brief overview of methods and how they deal with or help with these unique characteristics before discussing methods for MOHPO in detail in Section~\ref{sec:multi_objective_methods}.
Bayesian optimization (Section~\ref{ssec:model_based_optimization}) employs a probabilistic surrogate of the objective function and thus enables the user to train models with nearly only HPCs relevant to finding the optimal HPC. Also, since BO variants can return a batch of HPCs, models can be trained in parallel, further improving the runtime. Mixed and hierarchical search spaces can be treated with BO with special kernel functions\footnote{This is non-trivial and usually can often not be done out of the box} or by using a suitable surrogate, e.g., random forests. Evolutionary algorithms (Section~\ref{ssec:evolutionary_algorithms}), on the other hand, can not select HPCs as effectively as BO and thus usually need more proposals than BO; however, they propose HPCs naturally in batches and can handle mixed and hierarchical search spaces with ease because of the discrete nature of their proposal-generating operations. 
Multi-fidelity methods (Section~\ref{sec:multifid}) may be combined with an optimizer to further address the expensive nature of the problem as they improve resource allocation.\\ 
We generally assume that all objective functions are \textit{black-boxes}, i.e., analytical information about $c_i$ is not available, even though there are exceptions to this (e.g., number of parameters in some models). 
To record the evaluated hyperparameter configurations and their respective scores, we introduce the so-called \emph{archive} $\mathcal{A} = ((\lambdav^{(1)}, c(\lambdav^{(1)})), (\lambdav^{(2)}, c(\lambdav^{(2)})), \dots)$, with $\mathcal{A}^{[t+1]} = \mathcal{A}^{[t]} \cup (\lambdav^+, c(\lambdav^+))$ if a single configuration is presented by an algorithm that iteratively proposes hyperparameter configurations.


\subsection{Multi-objective machine learning}

A concept closely related but different to MOHPO is multi-objective machine learning. 
To understand the difference, it is important to differentiate between first level \textit{model parameters} (e.g., weights of a neural network or learned decision rules) and second order \textit{hyperparameters} (e.g., neural network architectures and optimizers)~\citep{bischl2021hyperparameter}. 
Model parameters are fixed by the ML algorithm at training time in accordance to one or multiple metrics, whereas hyperparameters are chosen by the ML practitioner before training and influence the behavior of the learning algorithm and the structure of its associated hypothesis space.
We define multi-objective ML methods as those that focus on learning first level parameters (sometimes together with second level hyperparameters). 
Our work, in contrast, concentrates on \textit{hyperparameter optimization} for ML algorithms, i.e., tuning \textit{second} level hyperparameters.
While the search space is generally smaller for hyperparameter optimization, the problem tends to be more expensive as multiple evaluations of the ML algorithm are required.
Feature selection is a topic that borders MOHPO and multi-objective ML and is often handled in a multi-objective manner~\citep{binder2020mosmafs, mierswa2006information, banerjee2006feature_selection_rough}. 
We consider feature selection as closely related to HPO and will therefore dedicate large portions of Section~\ref{ssec:sparseness} to this topic.
However, it should be noted that feature selection is often grouped with multi-objective ML, as e.g., in~\citet{jin2006multi}.

\section{Foundations of multi-objective optimization}\label{sec:foundations}
This section covers some of the fundamental background to multi-objective optimization. Readers familiar with the topic can directly jump to the discussion of solution algorithms in Section~\ref{sec:multi_objective_methods} or even Section~\ref{sec:applications}, where applications are presented.
\subsection{Objectives and constraints}\label{ssec:objectives_constraints}

In the context of this paper, objectives refer to the  evaluation criteria of the ML model $c_1: \LamS \to \R, \dots, c_m: \LamS \to \R$ with $m \in \N$. While oftentimes we simply aim to minimize these objectives, in real-world applications, we may well face a constrained HPO problem of the form:
\begin{equation}
\label{mohpo_problem}
\begin{aligned}
& \underset{\lambdab \in \LamS}{\argmin}
& & c(\lambdab) \\
& \text{subject to}
&& k_1(\lambdab) = 0, \dots, k_n(\lambdab) = 0 \; &&(\text{equality constraints}), \\
&&& \hat{k}_{1}(\lambdab) \geq 0, \dots, \hat{k}_{\hat{n}}(\lambdab) \geq 0\; &&(\text{inequality constraints}),
\end{aligned}
\end{equation}
where $c: \LamS \to \R^m$ as before and $k_1, \dots k_n$ (equality constraint functions) and $\hat{k}_{1}, \dots, \hat{k}_{\hat{n}}$ (inequality constraint functions) are functions on $\LamS$ with one dimensional, real output.
While equality constraints are not as prevalent in the context of (MO)HPO, they sometimes are included; a very simple example is requiring a certain number of clusters in the final result of a clustering algorithm~\citep{yang2020hyperparameter}.
Inequality constraints are often relevant to (MO)HPO problems:
In many applications, predictive performance should not cross a certain threshold, energy consumption of an ML model should be under a given amount or if fairness with respect to two subpopulations satisfies a specific value, further improvement is not required.
It is the task of an ML practitioner to translate a real-world problem into an ML task - and therefore objectives and constraints - to measure the quality and feasibility of a given model.
Depending on the use case it has to be carefully considered whether to frame a requirement to an ML model as an objective or a constraint.
For example, is it important to have a model as memory-efficient as possible or is memory requirement limited by a hard constraint? 
Finding high performing and efficient deep learning architectures has emerged as a prominent task recently, coining the term hardware-aware NAS (HW-NAS)~\citep{ijcai2021p592}.
Successful approaches exist that frame the HW-NAS problem as a constrained optimization problem~\citep{cai2019proxylessnas, tan2019mnasnet} or a MOO problem~\citep{elsken2018efficient,lu2019nsga}.
It should be noted that constrained optimization comes with its own set of challenges that are orthogonal to the multi-objective aspect we focus on in this paper. 
We therefore exclude constrained optimization from our work - only mentioning it in absolutely crucial parts and giving helpful references to the reader when appropriate.

\subsection{Pareto optimality}\label{ssec:optimality}
A vector of multiple objectives $c \in \R^m$ \emph{(Pareto-)dominates} another vector of multiple objectives $c^\prime,$ written as\footnote{In some literature, the direction of the domination relationship is reversed, i.e., they write $c^\prime  \prec c$ if $c$ dominates $c^\prime$. We choose our notation because it naturally fits the minimization perspective taken in this paper.} 
$c  \prec c^\prime$, if and only if
\begin{align}
  \label{eq:pareto-dominance-1}
  \begin{split}
  \forall i \in \left\{1, ..., m\right\}&: c_i \le c^\prime_i\, \land \\
  \exists j \in \left\{1, ..., m\right\}&: c_j < c^\prime_j.
  \end{split}
\end{align}
Analogously, we say a HPC $\lambdab \in \LamS$ \emph{(Pareto-)dominates} another configuration $\lambdab^\prime$, if and only if $c(\lambdab) \prec c(\lambdab^\prime).$
In other words: $\lambdab$ dominates $\lambdab^\prime$, if and only if there is no criterion $c_i$ in which $\lambdab^\prime$ is superior to $\lambdab$, and at least one criterion $c_j$ in which $\lambdab$ is strictly better.

\begin{figure}[ht]
    \centering
    \includegraphics[width = 0.9\textwidth]{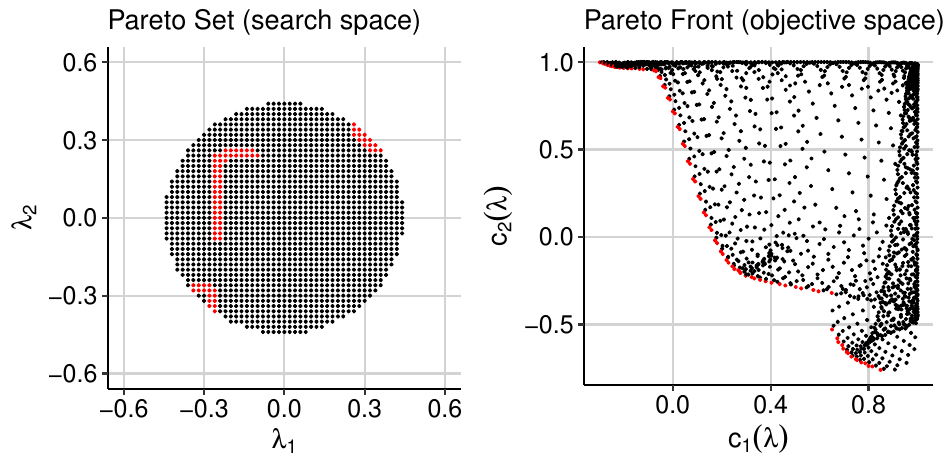}%
    \caption{Illustration for a two-dimensional MOO problem with two objectives $c_1$ and $c_2$. The left plot shows the search space $\LamS$, and the right plot shows the objective space $\R^2$. Non-dominated configurations - the estimated Pareto set (left) - and their mapping to the co-domain - the estimated Pareto front (right) - are highlighted.}
    \Description{Left plot: Two-Dimensional spherical search space where the Pareto optimal sets are unconnected. Right plot: Display of the 2 dimensional associated Pareto front.}
    \label{fig:example_front}
\end{figure}

We say $c$ \emph{weakly dominates} $c^\prime$, written as $c \preceq c^\prime$, if and only if $c \prec c^\prime$ or $\forall i \in \{1, \dots, m\}\;  c_i = c^\prime_i$. Analogously, if $c(\lambdab) \preceq c(\lambdab^\prime)$ we say the HPC $\lambdab$ weakly dominates $\lambdab^\prime.$
A configuration $\lambdab^\ast$ is called \emph{non-dominated} or \emph{(Pareto) optimal} if and only if there is no other $\lambdab \in \LamS$ that dominates $\lambdab^\ast$. 
Pareto dominance defines only a partial order over $\R^m$, i.e., two vectors of multiple objectives $c$ and $c^\prime$ can also be incomparable. This situation arises if 
there exist $i, j \in \{1, \dots, m\}$ for which $c_i < c^\prime_i$ but also $c^\prime_j < c_j$. 
Hence, in contrast to single-objective optimization, there is in general no unique single best solution $\lambdab^\ast$, but a set of Pareto optimal solutions that are pairwise incomparable with regard to $\prec$. 
This set of solutions is referred to as the \emph{Pareto (optimal) set} 
and defined as
\begin{eqnarray}
\mathcal{P} := \left\{\lambdab \in \LamS ~|~ \not \exists~ \lambdab^\prime \in \LamS \text{ s.t. } \lambdab^\prime \prec \lambdab\right\}.
\end{eqnarray}

The image of $\mathcal{P}$ under $c$, written as $c\left(\mathcal{P}\right)$, is referred to as the Pareto front (see Figure~\ref{fig:example_front}). 
The goal of a multi-objective optimizer that solves~(\ref{eq:multi_obj}) is not to find a single best configuration $\lambdab^\ast$, but rather a set of configurations $\mathcal{\hat P}$ that approximates the Pareto set $\mathcal{P}$ well.


\subsection{Evaluation}\label{ssec:evaluation}
\label{sec:evaluation}
The result of a multi-objective algorithm is $\hat{\mathcal{P}}$, the set of points of the estimated Pareto front. 
In order to evaluate this set or compare it to other sets, one must define what it means for a Pareto front to be better than another. 
Usually, this comparison is quantitatively based on so-called quality indicators.  

\subsubsection{Comparing solution sets}
\label{sec:compp_sol}

Let $\mathcal{S}_1$ and $\mathcal{S}_2$ be two non-dominated solution sets - i.e., within each set, no configuration is dominated by another configuration. 
The associated approximated Pareto fronts are denoted by $\hat{\mathcal{P}}_{\mathcal{S}_1}$ and $\hat{\mathcal{P}}_{\mathcal{S}_2}$ respectively.
According to~\citet{zitzler2003performance}, $\hat{\mathcal{P}}_{\mathcal{S}_1}$ is said to \textit{weakly dominate} $\hat{\mathcal{P}}_{\mathcal{S}_2}$, denoted as $\hat{\mathcal{P}}_{\mathcal{S}_1} \preceq \hat{\mathcal{P}}_{\mathcal{S}_2}$, if for every solution $\lambdab_2\in \mathcal{S}_2$ there is at least one solution $\lambdab_1\in\mathcal{S}_1$ which weakly dominates $\lambdab_2$. 
$\hat{\mathcal{P}}_{\mathcal{S}_1}$ is furthermore said to be \textit{better} than $\hat{\mathcal{P}}_{\mathcal{S}_2}$, denoted\footnote{In some literature, the direction of the \textit{better}-relationship is reversed, i.e., they write $\hat{\mathcal{P}}_{\mathcal{S}_1}  \triangleright \hat{\mathcal{P}}_{\mathcal{S}_2}$ if $\hat{\mathcal{P}}_{\mathcal{S}_1}$ is better than $\hat{\mathcal{P}}_{\mathcal{S}_2}$. We choose our notation because it naturally fits the minimization perspective taken in this paper.} as $\hat{\mathcal{P}}_{\mathcal{S}_1} \triangleleft \hat{\mathcal{P}}_{\mathcal{S}_2}$, if
$\hat{\mathcal{P}}_{\mathcal{S}_1} \preceq \hat{\mathcal{P}}_{\mathcal{S}_2}$, but not every solution of $\hat{\mathcal{P}}_{\mathcal{S}_1}$ is weakly dominated by any solution in $\hat{\mathcal{P}}_{\mathcal{S}_2}$, i.e.,  $\hat{\mathcal{P}}_{\mathcal{S}_2} \not\preceq \hat{\mathcal{P}}_{\mathcal{S}_1}$. This represents the weakest form of superiority between two approximations of the Pareto front. 
Note that these order relationships defined for $\hat{\mathcal{P}}_{\mathcal{S}_1}$ and $\hat{\mathcal{P}}_{\mathcal{S}_2}$ can directly be transferred to the associated solution sets $\mathcal{S}_1$ and $\mathcal{S}_2$.
How well a single solution set represents the Pareto front can be divided into four qualities~\citep{li2019quality}:
\begin{description}
    \item[Convergence] The proximity to the true Pareto front
    \item[Spread] The coverage of the Pareto front
    \item[Uniformity] The evenness of the distribution of the solutions
    \item[Cardinality] The number of solutions
\end{description}
The combination of spread and uniformity is also referred to as \textit{diversity}.
One way of comparing these qualities is to visualize the solution sets. 
For bi-objective optimization problems, this can be straightforwardly done. 
However, for a higher number of objectives, the visualization and decision-making process based on this visualization can become substantially more challenging. 
\citet{tusar2014visualization} offer a review of existing visualization methods.

\subsubsection{Quality indicators}
\label{ssec:qual_ind}
An objective measurement of the quantitative difference between solution sets is clearly desirable for comparing algorithms. 
Therefore, many quality indicators $I$ - which map the approximation of the Pareto front to a real number representing the quality of a set of solutions - were proposed. 
An extensive overview of these indicators can be found in~\citet{li2019quality}. 
Quality indicators that focus on all four qualities listed above can be divided into \textit{distance-based}, which require the knowledge of the true Pareto front or a suitable approximation of it, and \textit{volume-based}, which measure the volume between the approximated Pareto Front and a method-specific point.
Common distance-based quality indicators are the inverted generational distance~\citep{coello2004Study}, Dist2~\citep{czyzzak1998pareto} and the $\epsilon$-indicator~\citep{zitzler2003performance}. 
Common volume-based indicators are hypervolume indicator~\citep{zitzler1998multiobjective}, the R class of indicators~\citep{Hansen1998EvaluatingTQ}, and the integrated preference functional~\citep{carlye2003Evaluating}. 
The most popular quality indicator is the \emph{hypervolume}~\citep{zitzler1998multiobjective}, also called $\mathcal{S}$-metric, since it does not require any prior knowledge of the Pareto front.
The hypervolume, $\operatorname{HV}$, of an approximation of the Pareto front $\hat{\mathcal{P}}_\mathcal{S}$ can be defined as the union of the dominated hypercubes $\text{domHC}_{\bm{r}}$ of all solution points $\lambdab \in \mathcal{S}$ regarding a reference point $\bm{r}$, i.e.,
\[ \label{eq:hypervolume}
  \operatorname{HV}_{\bm{r}}(\mathcal{S}) := \mu\left(\bigcup_{\lambdab \in \mathcal{S}}\text{domHC}_{\bm{r}}(\lambdab)\right),
\]
where $\mu$ is the Lebesgue measure and the dominated hypercube
\[
  \text{domHC}_{\bm{r}}(\lambdab) := \{\bm{u} \in \R^m~|~ c_i(\lambdab) \leq \bm{u}_i \leq \bm{r}_i\;\forall i \in \{1, \dots, m\}\}.
\]

\begin{figure}[ht]
    \centering
    \includegraphics[width = 0.6\textwidth]{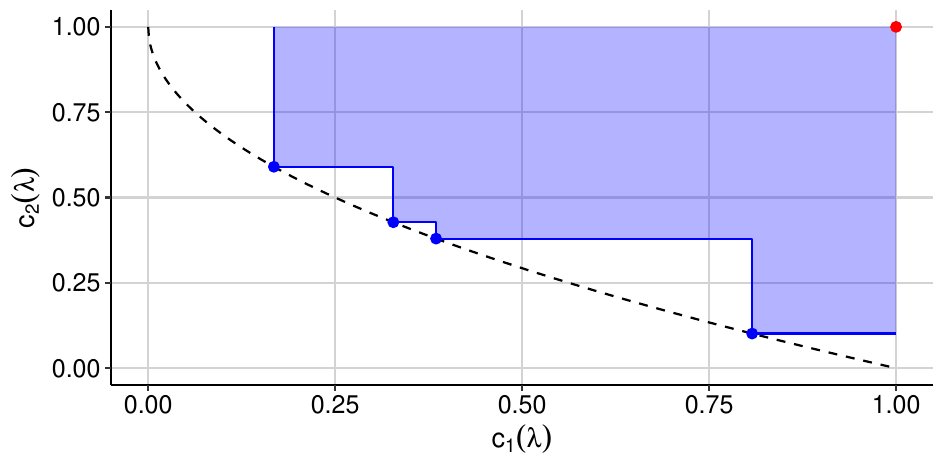}
    \caption{ The plot shows the hypervolume indicator (the area of the blue shaded region) regarding the reference point $(1,1)$ (marked in red). 
    }
    \label{fig:qual_ind}
    \Description{Display of the dominated hypervolume for a bi-objective optimization problem regarding four undominated candidates}
\end{figure}

$\operatorname{HV}$ is illustrated in Figure~\ref{fig:qual_ind}.
The hypervolume indicator is \textit{strictly Pareto compliant}~\citep{zitzler2006hypervolume}, i.e., 
for all solution sets $\hat{\mathcal{P}}_{\mathcal{S}_1}$ and $\hat{\mathcal{P}}_{\mathcal{S}_2}$, it holds that
\[ 
  \mathcal{S}_2 \triangleright \mathcal{S}_1 \Rightarrow  \operatorname{HV}_{\bm{r}}(\mathcal{S}_2) <  \operatorname{HV}_{\bm{r}}(\mathcal{S}_1). 
\]
Hence for the true Pareto front, the $\operatorname{HV}$ reaches its maximum. 
In practice, the \emph{nadir point}, which is constructed from the worst objective values, or a point slightly beyond it, is often used as the reference point. For a guideline on how to set a reference point, see~\cite{Ishibuchi18referencepoint}.
Based on a quality indicator estimate, MOO strategies can be evaluated over different benchmark datasets by using appropriate statistical tests.
\citet{eftimov2018data} on the other hand propose to use a data-driven preference-based approach, which includes information about the influence of each quality indicator.

\subsubsection{A note on normalizing objectives}
\label{sssec:normalizing_obejctives}
As different objectives in MOHPO can live on different scales, maybe even with different orders of magnitude, this may bias evaluation of some quality indicators like hypervolume towards certain objectives.
While this issue is discussed in general MOO literature, in most MOHPO publications it is either completely ignored or dealt with in an ad-hoc form, without further justification. 
Transformations of the objective space are performed to deal with different orders of magnitude as e.g., in~\citet{hernandez2016predictive}, but the issue of normalization is seldom discussed in publications.
To make MOO more reliable and prevent bias towards an objective, it is generally recommended to normalize objectives to the interval $[0,1]$ ~\citep{grodzevich2006normalization, he2021survey, horn2015model}.
This should at least be done for analysis and comparison of algorithms~\citep{horn2016multi, horn2015model}
The importance of normalized objectives and various methods on how to implement them have been of interest within the multi-objective evolutionary computation community; even the particular effects on certain algorithms has been examined in detail~\citep{he2021survey, blank2019investigating, ishibuchi2017effect}.
For surrogate-assisted optimization methods, such as Bayesian optimization, normalization of objectives is often presented as a challenge~\citep{knowles2006parego, hernandez2016predictive}, but only few works exist that examine its effects \citep{wang2021investigating, wang2021adjusting}.
In general, if limits of the objectives are known \emph{a priori}, the \emph{nadir} and \emph{ideal point} can be used to scale objectives.
While the \emph{ideal point} may be found by solving a single-objective optimization problem for each objective, this is not necessarily practical for expensive optimization problems.
The \emph{nadir point} is harder to determine or even estimate, as it requires knowledge about the Pareto front~\citep{he2021survey}.
The estimation of \emph{ideal} and especially \emph{nadir point} has spawned a number of dedicated publications~\citep{deb2009review, wang2017nadir}.
A simple and often effective method to estimate limits is to use previously evaluated points: The \emph{ideal} and \emph{nadir point} can then be estimated either through the non-dominated solutions or the entire archive~\citep{blank2019investigating}.
In iterative optimization procedures, these estimates can be updated after new evaluations have become available, which is often adopted in practice~\citep{pfisterer_yahpo}.
The exact effect and challenges of objective normalization depend on the chosen MOO method: While some MOO methods are more robust towards not normalizing objectives, for others it is an integral part of the optimization procedure. We will provide comments for specific methods when appropriate in Section~\ref{sec:multi_objective_methods} (e.g., if a method does not require normalization or if it is an essential part of a certain algorithm).

\subsubsection{A note on multimodal multi-objective optimization}
Multimodality is a relevant issue in various application domains of MOO~\citep{tanabe2019review, li2016seeking}.
If e.g., several candidates are in close proximity to each other in objective space, but represent a very heterogeneous set and high diversity in decision space, it may be desirable to identify all of them during optimization~\citep{tanabe2019review}:
If parts of the decision space become infeasible, 
the decision maker can then utilize alternatives that represent very similar trade-offs.
While the relevance to his in MOHPO is debatable,~\citet{grimme2021peeking} present a second issue - multimodality as an obstacle for optimization in identifying global optima (i.e., not getting ``trapped'' in local minima), which is an understudied problem with great potential~\citep{grimme2021peeking}. 
It may be of interest to adapt an optimization algorithm to take diversity in decision space into consideration as is sometimes done in single-objective HPO through e.g., a multi-objective infill criterion in Bayesian optimization~\citep{bischl2014moimbo}.
When it comes to measuring solution set quality with respect to identifying all global optima, this is very much still an open question:
No commonly accepted quality indicators exist and those approaches previously proposed lack theoretical foundation~\citep{grimme2021peeking}.

\section{Multi-objective optimization methods}
\label{sec:multi_objective_methods}

This work in general and the following section in particular, focus on \textit{a posteriori} methods, i.e., those that return a set of configurations $\hat{\mathcal{P}}$ that tries to approximate the true Pareto set $\mathcal{P}$ as well as possible. 
Nevertheless, we will start by briefly discussing some \emph{a priori} methods, i.e., those that will only return one configuration depending on preferences set before optimization, as they are a benchmark and basis for various a posteriori methods.
While a multitude of very specific methods for MOO exist across various domains, we try to mainly focus on those that are actually applied to MOHPO.
Further methods can be found within the field of algorithm configuration, where multi-objective problems are also of interest~\citep{blot2016mo, zhang2013s}.

\subsection{Scalarization approaches}\label{sssec:scalarization}

Scalarization transforms a multi-objective goal into a single-objective one, i.e., it is a function $s: \R^m \times \mathcal{T} \rightarrow \R$ that maps $m$ criteria to a single criterion to be optimized, configured by scalarization hyperparameters $\alpha \in \mathcal{T}$. 
Having only one objective often simplifies the optimization problem~\citep{miettinen2012nonlinear}. 
However, there are two main drawbacks to using scalarization for MOO~\citep{jin2008pareto}: Firstly, the scalarization hyperparameters $\alpha$ must be chosen sensibly, such that the single-objective represents the desired relationship between the multiple criteria -- which is not trivial, especially without extensive prior knowledge of the optimization problem. Secondly, one single solution can not adequately represent a multi-objective problem with conflicting objectives.
We will outline three popular scalarization techniques:

\paragraph{Weighted sum approach}
One of the most well-known scalarization techniques, where one looks for the optimal solution to: 
\begin{equation}
\label{eq:weighted_sum}
    \min_{\lambdab \in \LamS} \sum^{m}_{i=1}\alpha_i c_i(\lambdab).
\end{equation}
It can be shown~\citep{ehrgott2005multicriteria} that for a solution $ \lambdab^*$ of (\ref{eq:weighted_sum}), it holds that if 
\begin{equation}
     \alpha_i \geq 0, \quad i=1,\dots,m \Rightarrow \lambdab^* \textrm{ is non-dominated in } \LamS.
\end{equation}
Additionally, it holds for convex $\LamS$ and convex functions $c_i, i=1,\dots,m$ that for every non-dominated solution $\lambdab^*$ there exist $\alpha_i \geq 0, \; i=1,\dots,m$, such that $\lambdab^*$ is the respective solution of (\ref{eq:weighted_sum}); however there is no combination of weights that would result in Pareto-optimal solutions that lie on a non-convex part of the Pareto frontier~\citep{ehrgott2005multicriteria}.

\paragraph{Tchebycheff approach}
The \emph{weighted Tchebycheff problem} is formulated as:
\begin{equation}
\label{eq:weigthed_tchebycheff}
\min_{\lambdab \in \LamS} \; \max_{i = 1, \dots, m} [\alpha_i | c_i(\lambdab) - z^*_i |],
\end{equation}
where $z^*_i = min_{\lambdab \in \LamS} c_i(\lambdab) \, for \, i = 1, \dots, m$ defines the (ideal) reference point.
For every optimal solution $\lambdab^*$ there exists one combination of weights $\hat{\alpha}$, so that $\lambdab^*$ is the optimal solution to the optimization problem~(\ref{eq:weigthed_tchebycheff}) defined through $\hat{\alpha}$.

\paragraph{$\epsilon$-constraint approach}
As outlined in Section~\ref{ssec:objectives_constraints}, in some instances it might be feasible to formulate one or more of the objectives as constraints. 
A MOO problem can also be reduced to a single scalar optimization problem by turning all but one objective into constraints, as in the \emph{$\epsilon$-constraint method}: Given $m-1$ constants $(\epsilon_2,\dots, \epsilon_{m}) \in \mathbf{R}^{m-1}$,
\begin{eqnarray}
    \min_{\lambdab \in \LamS} c_1(\lambdab),\; \:  \text{subject to} \; c_2(\lambdab) \leq \epsilon_2,\dots,c_{m}(\lambdab) \leq \epsilon_{m}.
\end{eqnarray}
However, much like parameters in the weighted sum method, constraints must be sensibly chosen, which can prove to be challenging \textit{a priori} and without sufficient domain knowledge.\\
\\
These scalarization techniques are also used in the MOEA/D Evolutionary Algorithm as outlined in Section~\ref{sssec:prominent_moeas} or the BO technique ParEGO as outlined in Section~\ref{sssec:mo-mbo}, where several scalar optimization problems are created in place of the multi-objective one.

\subsection{Random and grid search} \label{sssec:random_and_grid_search}

Random and grid search are simple baseline algorithms for single-objective HPO~\citep{bischl2021hyperparameter}.
Random search is generally preferred, as it is an anytime algorithm and scales better in the case of low effective dimensionality of an HPO problem with low-impact parameters~\citep{bergstra2012random}; it often provides a 
surprisingly competitive baseline.
Modifying random and grid search for MOO is trivial: as all points are independently spawned and  evaluated, one simply returns all non-dominated solutions from the archive.
Random and grid search can serve as reasonable baselines when introducing more sophisticated optimization methods - similar to the single-objective case. This procedure is adopted in a number of MOHPO works~\citep{parsa2020bayesian, schmucker2020multi_obj_multi_fid, iqbal2020flexibo}

\subsection{Evolutionary algorithms}
\label{ssec:evolutionary_algorithms}
Evolutionary algorithms (EAs) are general black-box optimization heuristics inspired by principles of natural evolution. This section provides a brief introduction, followed by some prominent examples of multi-objective EAs. EAs haven been used in HPO and AutoML applications, such as in the popular AutoML framework TPOT~\citep{Olson2016EvoBio}.
While they generally require more evaluations than e.g., Bayesian optimization and thus may not seem like an intuitive choice for expensive optimization problems, they are very adept at handling mixed, hierarchical or otherwise awkward search spaces.
Swarm Intelligence methods are related to and share many of the advantages and disadvantages of EAs.
They can be utilized for multi-objective optimization~\citep{fieldsend2009swarm} and have been examined for MOHPO and feature selection in select works (e.g.,~\citet{xue2012particle},~\citet{bacanin2020optimizing}).

\subsubsection{Fundamentals of evolutionary algorithms}

\begin{figure}
    \centering
    \includegraphics[width=0.4\textwidth]{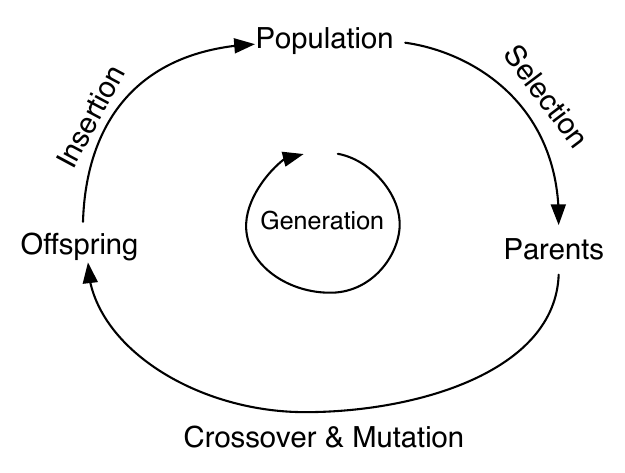}
    \caption{Basic loop of an evolutionary algorithm.}
    \label{fig:EA}
    \Description{Parents are selected from a population. Offsprings are created from the parents via crossover & mutation. These offsprings are inserted into the old population from which new parents can be selected again.}
\end{figure}

Evolutionary algorithms (EAs) are population-based, randomized meta-heuristics, inspired by principles of natural evolution. Historically, different variants have been independently developed  such as genetic algorithms, evolution strategies and evolutionary programming. These are now generally subsumed under the term ``evolutionary algorithm''.
EAs start by (often randomly) initializing a \emph{population} of solutions and evaluating them. Then, in every iteration (often called \emph{generation}), the better solutions are (often probabilistically) selected as \emph{parents} and used to generate new solutions, so-called \emph{offspring}. The two main operators to generate new solutions are \emph{crossover}, which tries to recombine information from two parents into an offspring, and \emph{mutation}, which randomly perturbs a solution. The resulting offspring solutions are then evaluated and inserted into the population. Since the population size is kept constant, some solutions have to be removed (\emph{survival selection}), and once again usually the better solutions are chosen to survive with a higher probability. If the stopping criterion is reached, the best encountered solution is returned, otherwise the next iteration starts.
A schematic overview of a generation can be found in Figure~\ref{fig:EA}.
EAs are popular optimization methods for the following reasons:
(i) no specific domain knowledge is necessary (at least for baseline results)~\citep{abraham2006evolutionary}, 
(ii) ease of implementation~\citep{abraham2006evolutionary}, 
(iii) low likelihood of becoming trapped in local minima~\citep{vachhani2015survey},
(iv) general robustness and flexibility~\citep{guliashki2008survey}, 
 and (v) straightforward parallelization~\citep{abraham2006evolutionary}. 
They are widely employed in practice and known for successfully dealing with complex problems and complex search spaces where other optimizers may fail~\citep{he2020automl}. 
Furthermore, because they are population-based, EAs are particularly well suited for multi-objective optimization, as they can simultaneously search for a set of solutions that approximate the Pareto front. 

\subsubsection{Multi-objective evolutionary algorithms (MOEAs)}

When applying EAs to a multi-objective problem, the only component that needs changing is the selection step (selecting parents and selecting survivors to the next generation). Whereas in single-objective optimization the objective function can be used to rank individuals and select the better ones, in multi-objective optimization there are many Pareto-optimal solutions with different trade-offs between the objectives, and we are interested in 
finding a good approximation of the Pareto front. 
Recent works categorize MOEAs into three classes~\citep{emmerich2018tutorial}:
\begin{itemize}
    \item Pareto dominance-based algorithms use two-levels for ranking: On the first level, Pareto dominance is used for a coarse ranking (usually non-dominated sorting, see below), while on the second level usually some sort of diversity measure is used to refine the ranking of the first level.
    \item Decomposition-based algorithms utilize scalarization (see Section~\ref{sssec:scalarization}) to decompose the original problem into a number of single-objective subproblems with different parametrizations, which are then solved simultaneously.
    \item Indicator-based algorithms use only a single metric, such as the hypervolume indicator, and selection is governed by a solution's marginal contribution to the indicator.
\end{itemize}

In the following, we will provide a prominent example for each category.

\subsubsection{Prominent MOEAs}
\label{sssec:prominent_moeas}

\paragraph{NSGA-II (Pareto dominance-based)}
The \emph{non-dominated sorting genetic algorithm} (NSGA-II)~\citep{deb2002nsgaII} is still one of the most popular MOEAs. 
In many benchmark studies in the field, it serves as a popular baseline~\citep{vachhani2015survey,parsa2020bayesian}.
Being Pareto dominance based, it first uses \emph{non-dominated sorting} to obtain an initial coarse ranking of the population. This iteratively determines the non-dominated solutions, assigns them to the next available class, and removes them from consideration. Among the solutions in each obtained class, the extreme solutions (best in each objective) are ranked highest, and the remaining solutions are ranked according to the \emph{crowding distance}, the sum of differences between an individual’s left and right neighbor, in each objective, where large crowding distances are preferred. 
While this works very well for problems with two objectives, it breaks down in case of a larger number of objectives, as the non-dominated sorting becomes less discriminative, and the left and right neighbor in each objective are often different solutions. NSGA-III~\citep{deb2014nasgaIIIpart1,deb2014nasgaIIIpart2} has been developed as an alternative for problems with a higher number of objectives, but shares many similarities with decomposition-based algorithms. 

\paragraph{MOEA/D (Decomposition-based)}
The \emph{Multi-objective Evolutionary Algorithm based on Decomposition} (MOEA/D) decomposes the multi-objective problem into a finite number $N$ of scalar optimization problems that are then optimized simultaneously~\citep{zhang2007moea}. 
Each single optimization problem usually uses a Tchebycheff scalarization, see Section~\ref{sssec:scalarization}.
In theory, each solution to such a scalar problem should be a point on the Pareto front of the original multi-objective problem.
The distribution of solutions on the Pareto front is thus governed by the set of scalarizations chosen, and 
it is challenging to identify suitable scalarizations without a good knowledge of the Pareto frontier.
Rather than solving the different scalarized problems independently, the idea is to solve them simultaneously, and allow the different search processes to inﬂuence each other. In a nutshell, the population comprises of the best solution found so far for each of the sub-problems. In every generation, a new offspring is created for each sub-problem by randomly selecting two parents from the sub-problem’s neighborhood, performing crossover and mutation, and re-inserting the individual into the population. The new individual replaces all individuals in the population for which it is better with respect to the corresponding sub-problem. In effect, this means mating is restricted to among individuals from the same region of the non-dominated frontier, and diversity in the population is maintained implicitly by the deﬁnition of the different sub-problems.

\paragraph{SMS-EMOA (Indicator-based)}

The \emph{S metric selection evolutionary multi-objective optimization algorithm} (SMS-EMOA)~\citep{emmerich2005emo} also uses the non-dominated sorting algorithm from NSGA-II for an initial coarse ranking, but then 
uses an individual's marginal hypervolume contribution as a secondary criterion. The marginal hypervolume of an individual $i$ is the difference in hypervolume between the population $\mathcal{R}$ including individual $i$, and excluding individual $i$: 
\begin{equation}
    \Delta_{HV}(i, \mathcal{R}) :=  HV(\mathcal{R}) - HV(\mathcal{R} \setminus i)
\end{equation}
Different from the other two algorithms above, SMS-EMOA only produces  one offspring per generation, adds it to the population and then discards the worst solution based on the ranking just described.

For more details on the above MOEAs, different MOEAs, and example applications, please refer to~\citet{coello2007evolutionary, abraham2006evolutionary, deb2001multi, branke2008multiobjective}.
The main disadvantage of many evolutionary algorithms is their relatively slow convergence (compared to other optimization methods) and the need for many evaluations; for more computationally expensive ML problems, a multi-objective HPO can become very costly when tackled  with these methods~\citep{deb2001multi}.
To alleviate this problems, MOEAs have been combined with surrogate-modelling techniques (e.g.,~\cite{lu2020nsganetv2}) or gradient-based local search (e.g.,~\cite{Lara10LocalSearch}).
EAs can further be found in several multi-objective AutoML solutions, such as TPOT~\citep{olson2016tpot} or FEDOT~\citep{polonskaia2021multi}.



\subsubsection{Relevant software and implementations}
Two very established packages which offer EMOAs are 
PlatEMO (MATLAB)\footnote{\url{https://github.com/BIMK/PlatEMO}} and pymoo (Python)\footnote{\url{https://github.com/anyoptimization/pymoo}}.
They both offer a wide range of EMOAs and can be generally recommended.
The mle-hyperopt (Python)\footnote{\url{https://github.com/mle-infrastructure/mle-hyperopt}} package offers directly MOHPO via NSGA-II using internally the nevergrad (Python)\footnote{\url{https://github.com/facebookresearch/nevergrad}} package.

\subsection{Model-based optimization}
\label{ssec:model_based_optimization}


\subsubsection{Bayesian Optimization} 
\label{sssec:model_based_optimization}
In the following, the basic concepts of Bayesian optimization (BO) are shown. For more detailed information, see~\citep{bischl2021hyperparameter}. BO is quite sample efficient compared to other optimization techniques~\citep{hernandez2016predictive}, which makes it a good choice for the expensive black-box problems presented by many machine learning models 
and specifically for HPO~\citep{jones1998efficient,hutter2011sequential,snoek_practical_2012}.
BO is an iterative algorithm with the key strategy of modelling the mapping $\lambdav~\mapsto~c(\lambdav)$ based on observed performance values found in the archive $\archive$ via (non-linear) regression. This approximating model is called a \textit{surrogate model}. While GPs are arguably the most widely used surrogate models for BO they have a hard time with mixed and hierarchical search spaces; in MOHPO they are often replaced by e.g., random forests~\citep{horn2016multi}. 
BO starts on an archive $\archive$ of evaluated configurations, typically sampled through Latin Hypercube Sampling or Sobol Sampling to ensure the diversity of solutions  \citep{bossek_initial_2020}.
%
BO then uses the archive to fit the surrogate model, which for each $\lambdav$ predicts both a performance $\chlam$ as well as an estimate of prediction uncertainty $\shlam$.
Based on the predictive distribution, BO computes a cheap-to-evaluate acquisition function $\ulam$ that encodes a trade-off between \textit{exploitation} and \textit{exploration} where exploitation favours solutions with high predicted performance, while exploration favours solutions with high uncertainty of the surrogate model because the surrounding area has not yet been explored sufficiently.
The acquisition function $\ulam$ is optimized in order to identify a new candidate $\lambdav^+$ for evaluation. The true objective value $c(\lambdav^+)$ of this proposed HPC $\lambdav^+$ is finally evaluated and added to the archive $\archive$. The surrogate model is updated, and BO iterates until a predefined termination criterion is reached.

\paragraph{Simple acquisition functions}
A very popular acquisition function is the \emph{expected improvement} (EI)~\citep{jones1998efficient}) over the best solution found so far.
A further, very simple acquisition function is the \emph{lower confidence bound} (LCB)~\citep{jones2001taxonomy}. 
The LCB treats local uncertainty as an additive bonus at each $\lambdav$ to enforce exploration, which can be controlled with a control parameter $\kappa$. 

\subsubsection{Multi-objective Bayesian Optimization} 
\label{sssec:mo-mbo}

The previously presented BO framework can be extended to optimize a set of possibly conflicting objectives simultaneously.
In particular, the multi-objective extensions to the BO framework (MO-BO) can be divided into those which transform the MO problem into a single objective problem by scalarization and those which do not.
Some MO-BO methods utilize one or more scalarizations of the MOO problem to fit surrogates in order to approximate the Pareto front.
Algorithms that do not use scalarization of the outcomes instead often train independent surrogates for each output dimension.
By having a prediction for each output dimension, we can either obtain an acquisition function for each dimension and use a multi-objective optimizer to obtain a set of promising configurations, or we can build an acquisition function that aggregates the predictions for each dimension into a single-objective acquisition function. 
In the following, we discuss some of these approaches in more detail.

\paragraph{Scalarization and ParEGO} 
ParEGO is a scalarization-based extension of BO to MOO problems proposed by~\citet{knowles2006parego}.
First, one must determine a set of scalarization weights that will be used throughout all BO-iterations. 
This set of weights should ensure that the Pareto front is explored evenly and is created
using the following rule:
\begin{eqnarray}
\label{eq:parego_weights}
\left\{\bm{\alpha} = (\alpha_1, \alpha_2, ..., \alpha_m) ~\middle|~ \sum_{j = 1}^m \alpha_j = 1 ~\wedge~ \alpha_j = \frac{l}{s}, l \in \{0, 1, ..., s\}\right\},     
\end{eqnarray}
which generates $\binom{s+m-1}{m-1}$ different weight vectors.
Second, the output space of each of our $m$ objectives is normalized to $[0,1]$.
While this is fairly simple for a number of metrics relevant to machine learning (e.g., accuracy, AUC) due to known limits, other metrics (e.g., prediction time, energy consumption) present with unknown bounds, in which case
the bounds of the respective objectives have to be estimated \citep{knowles2006parego} (see Section \ref{sssec:normalizing_obejctives}).
Finally, in each iteration of ParEGO, we create the scalarized objective using the so-called augmented Tchebycheff function~\citep{knowles2006parego}:
\begin{eqnarray}
    \label{eq:parego_scalarization}
    c_{\bm{\alpha}}(\lambdab) = \max\limits_{j \in \{1, \ldots, m\}} \left(\alpha_j c_j(\lambdab)\right) + \rho [\bm{\alpha} \cdot c(\lambdab)], 
\end{eqnarray}
where $\rho$ is a small positive constant and $\bm{\alpha}$ is a weight vector drawn uniformly from the set in~(\ref{eq:parego_weights}).
The second term in the augmented Tchebycheff function ensures to break ties between solutions with the same maximal objective value.
In each iteration, a surrogate model is fitted on the design with scalarized outcomes 
$\left\{\left(\lambdab^{(i)}, c_{\bm{\alpha}}(\lambdab^{(i)})\right)\right\}_i$, 
and the EI is optimized on this model to propose a new HPC. 
ParEGO can easily be extended to parallel batch proposals.
In order to propose a batch of size $q$,~\citet{horn2015model} suggest to sample $q$ different weight vectors per iteration in a stratified manner. 
New design points are proposed by fitting $q$ surrogates to the $q$ differently scalarized outcomes in the design and optimizing the acquisition function on each of the $q$ surrogate models in parallel.
One concern with ParEGO and scalarizing-based solutions in general is that the uniformly sampled weights do not necessarily result in the best distribution of nondominated points.
An advantage of ParEGO is that it can be easily adapted to focus the search on one objective by limiting the maximum weights of the others.

\paragraph{EHI} (or EHVI)~\citet{emmerich2006single} proposed to use the expected improvement over the S-metric (i.e., hypervolume) as an acquisition function for MO-BO. 
Here, a surrogate model for each objective is fitted individually. 
The EHI is then calculated as the expectation of the hypervolume improvement over the distribution of outcomes as predicted by the surrogate models.
A drawback of hypervolume-based BO is that a non-trivial multidimensional integral must be evaluated to calculate the EHI. 
It is possible to use Monte-Carlo-based approximations~\citep{emmerich2011hypervolume}, and the KMAC method~\citep{yang2019efficient} to efficiently calculate the EHI criterion with complexity $O(n \log n)$ in three dimensions and $O(n^{\left\lfloor m/2\right\rfloor})$ in $m$ dimensions. 
Additionally,~\citet{emmerich2016multicriteria} propose a more efficient means of calculating the EHI with a complexity of $O(n \log n)$ for $m=2$.
However, these methods are significantly more complex than the other presented MO-BO infill criteria.
To obtain batch-proposals,~\citet{yang_multipoint_2019} propose dividing the objective space into several sub-objective spaces and then search for the optimal solutions in each sub-objective space by using a truncated EHI.

\paragraph{SMS-EGO} 
The $\mathcal{S}$-Metric Selection-based Efficient Global Optimization (SMS-EGO) algorithm is another popular extension of MO-BO. 
This method was proposed by~\citet{ponweiser2008multiobjective} and extends the idea of the EHI by employing an infill criterion: 
In each BO iteration, SMS-EGO approximates each of the $m$ objectives with a separate surrogate model. 
For each objective, we compute the LCB and denote the resulting $m$-dimensional outcome with $\bm{u}_{\operatorname{LCB}}$.
The the acquisition function value of a configuration $\lambdab$ is derived from the increment of the dominated hypervolume when $\bm{u}_{\operatorname{LCB}}(\lambdab)$ is added to the current Pareto front approximation $\mathcal{\hat P}$:
\begin{align}
    \label{eq:acq_sms}
    u_{\operatorname{SMS}}(\lambdab) = \operatorname{HV}_{\bm{r}} \left( \mathcal{\hat P} \cup \bm{u}_{\operatorname{LCB}}(\lambdab)\right)  - \operatorname{HV}_{\bm{r}}(\mathcal{\hat P}) - p,
\end{align}
with a penalty $p$ and with the reference point $\bm{r}$ chosen as $\max(\mathcal{\hat P}) + \bm{1}_m$. 
If $\bm{u}_{\operatorname{LCB}}(\lambdab)$ is a dominated solution, $u_{\operatorname{SMS}}(\lambdab)$ would be zero without the penalty term $p$, making the optimization of the acquisition function more challenging.
Therefore, to guide the search towards non-dominated solutions in areas of dominated solutions, a penalty $p$ is added for each point that dominates the solution candidate.

\paragraph{Multi-EGO}
In each BO iteration, Multi-EGO approximates each of the $m$ objectives with a separate surrogate model from which a single-objective acquisition function is obtained.
\citet{jeong2005efficient} use Gaussian processes for the surrogates and EI as the individual acquisition functions.
The $m$ acquisition functions are optimized jointly as an MOO problem itself using a multi-objective genetic algorithm (in principal any MOEA can be used here), resulting in a set of candidates with non-dominated acquisition function values.
From this set, Multi-EGO then selects multiple points to be evaluated.
This naturally lends to parallelization, as there are always multiple proposals generated in each iteration.

\paragraph{MESMO and PESMO} 
\label{subp:mesmo_and_pesmo}

Common information-theoretic acquisition functions are multi-objective maximum entropy search (MESMO) and predictive entropy search (PESMO) proposed by~\citet{belakaria2019maxvalue} and~\citet{hernandez2016predictive} respectively.
Information theoretic multi-objective acquisition functions model each objective with an independent surrogate model. 
The resulting Pareto set is modelled as a random variable.
Hence, we can compute the entropy of the location of the Pareto set. 
The lower this entropy is, the more we know about the location of the Pareto set. 
These acquisition functions represent the expected reduction of entropy if a point is evaluated. 
Information theoretic acquisition functions are linear combinations of the expected entropy of the whole predictive distributions of every objective. 
Hence, the utility of every point represents a global measure of the uncertainty of the input space and not a local and heuristic measure of the particular point, as in other acquisition functions. Thus, in theory they should explore the search space more efficiently~\citep{hernandez2016predictive,belakaria2019maxvalue}.

\subsubsection{Relevant software and implementations}
\label{sssec:bo_software}
Table~\ref{tab:mobo_fw} compares BO frameworks with multi objective capabilities. We identified as suitable well-established frameworks \emph{Dragonfly}~\citep{kirthevasan2020dragonfly} (Python), \emph{HyperMapper}~\citep{nardi2019hypermapper} (Python, written in large parts with C++), \emph{Openbox}~\citep{li2021openbox} (Python), \emph{Ax}\footnote{\url{https://ax.dev/}} (Python), \emph{trieste}\footnote{\url{https://secondmind-labs.github.io/trieste/}} (Python), BoTorch~\citep{DBLP:conf/nips/BalandatKJDLWB20} (Python), Optuna~\citep{optuna_2019} (Python), GPflowOpt~\citep{GPflowOpt2017} (Python), mlr3mbo\footnote{\url{https://cran.r-project.org/package=mlr3mbo}} (R). We checked if they offer scalarization-based approaches (e.g., ParEGO), hypervolume-based approaches (e.g., EHVI, QNEHVI, SMS-EGO), or information-theoretic approaches, and if they can handle constraints and use computational resources in parallel.   

\begin{table}[ht]
    \centering
      \caption{BO frameworks which offer MO support}
    \footnotesize
    \tabcolsep=0.11cm
    \begin{tabular}{ccccccccccc}
    \toprule
        & Dragonfly & HyperMapper & OpenBox & Ax & trieste & BoTorch & GPflowOpt & mlr3mbo & Optuna \\
        \midrule
         Scal.-based & $\checkmark$ & $\checkmark$& $\checkmark$ & $\checkmark$& $\checkmark$& $\checkmark$& $\times$ & $\checkmark$ & $\checkmark$\\
         HV-based  & $\times$ & $\times$& $\checkmark$& $\checkmark$& $\checkmark$& $\checkmark$& $\checkmark$ & $\checkmark$ & $\times$\\
         Info-based &$\times$ & $\times$& $\checkmark$& $\checkmark$ & $\times$ & $\checkmark$ & $\times$ & $\times$ & $\times$\\
         Constraints &$\times$ & $\checkmark$& $\checkmark$& $\checkmark$& $\checkmark$ & $\checkmark$& $\times$ & $\times$ & $\checkmark$\\
         Parallel & $\checkmark$ & $\checkmark$ & $\checkmark$ & $\checkmark$ & $\checkmark$ & $\checkmark$ & $\times$ & $\checkmark$ & $\checkmark$\\
         \bottomrule
    \end{tabular}
   \label{tab:mobo_fw}
\end{table}

\subsection{Multi-fidelity optimization}
\label{sec:multifid}

Assuming the existence of cheaper approximate functions, multi-fidelity optimization is a popular choice for expensive-to-evaluate black-box optimization target functions in many application domains like aerodynamics~\citep{forrester2007multi} or industrial design~\citep{huang2006sequential}.
Multi-fidelity optimization has been established for single-objective hyperparameter tuning~\citep{original_hyperband} and has become a desirable mode of optimization especially for deep learning models, that are very costly to fully train and evaluate~\citep{original_hyperband,Li2020asha,falkner2018BOHB}.
\begin{figure}[ht]
    \centering
    \includegraphics[width=0.45\textwidth]{"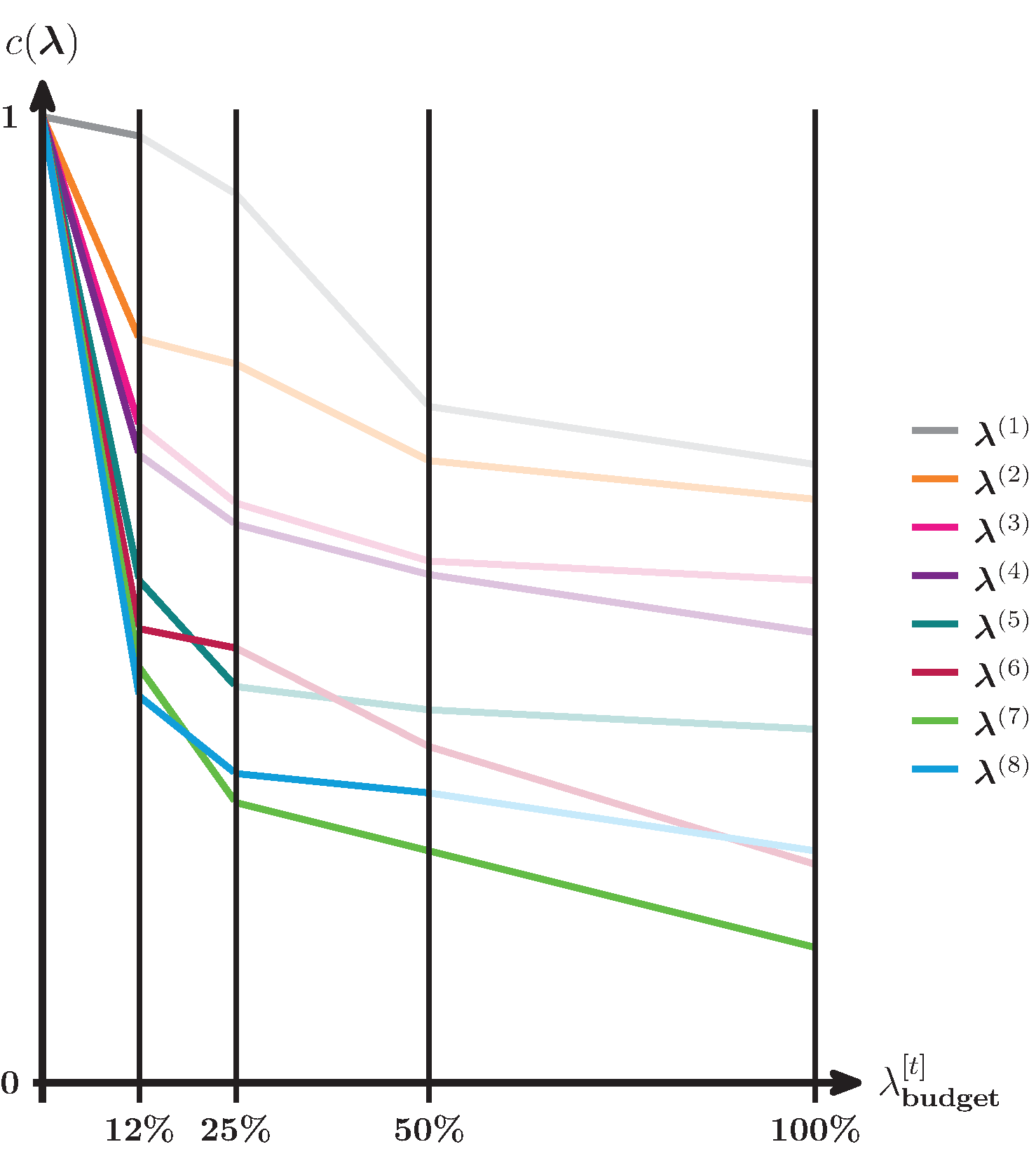"}
    \caption{Exemplary bracket run (figure inspired by~\citet{automl_book}). Faint lines represent future performance of HPCs that were discarded early.}
    \label{fig:hyperband}
    \Description{Curves of the objective values over the budget parameters are shown. Curves with high objective value at low budget tend to have also relatively high objective values at high budget. Successive halving successfully allocates more budget to the promising configurations.}
\end{figure}
Instead of optimizing configuration selection, as in model-based optimization, multi-fidelity methods for HPO aim at optimizing configuration evaluation by allocating resources in an efficient manner: Additional resources (e.g., additional epochs) are allocated to configurations that performed well with fewer resources, i.e., on a lower fidelity, and configurations that showed worse performance are discarded.
This idea is shown in Figure~\ref{fig:hyperband}.
Optimizing configuration evaluation is especially desirable in expensive optimization problems and while not all MOHPO problems fall in this category, many machine learning pipelines and especially deep learning models are quite expensive to evaluate.
When drawing configurations at random, multi-fidelity approaches can deal with the mixed and hierarchical search spaces of (MO)HPO nicely.
These methods have since been enhanced by the use of model-based optimization for drawing configurations instead of random sampling~\citep{falkner2018BOHB} and asynchronous execution~\citep{klein2020model}.
The same holds true in the multi-objective setting as recent works have shown~\citep{schmucker2020multi_obj_multi_fid, schmucker2021sh}.
Multi-fidelity extensions like Hyperband can be carried over to the MO setting by simply defining a suitable performance indicator to decide where increased resource allocation is desirable, which could be achieved in numerous ways.
Existing approaches have done this via scalarization using random weights~\citep{schmucker2020multi_obj_multi_fid} or non-dominated sorting~\citep{schmucker2021sh,salinas2021multi_obj_multi_fid_hw-nas}.
Taking multi-fidelity one step further, it can also be combined with BO, so configurations are no longer sampled randomly (as in traditional Hyperband), but via BO~\citep{falkner2018BOHB}.
Therefore, instead of enhancing random search, multi-fidelity methods are used to enhance BO in this case~\citep{falkner2018BOHB}.
The concept has recently been applied to MOO \citep{belakaria2020multi, irshad2021expected, chen2022towards}.
\citet{belakaria2020multi} introduce HPO of neural networks as one possible application for these methods and \citet{chen2022towards} introduce a multi-objective variant of BO Hyperband with a specific focus on applications in MOHPO.


\subsection{Further issues}

\subsubsection{Focusing optimization through user preferences}
\label{sssec:user_preferences}
A growing body of literature is exploring the integration of decision maker (DM) preferences into multi-objective optimization~\citep{Branke2016,wang2017mini}. Depending on when the user preferences are elicited relative to the optimization process, these methods are generally divided into \emph{a priori} (before optimization), \emph{progressive} (during optimization), and \emph{a posteriori} (after optimization) approaches~\citep{branke2005integrating}. 
The majority of literature on multi-objective optimization aims to find a good approximation of the entire Pareto front, providing the decision maker with a variety of alternatives to choose from \emph{after} optimization, so falls into the \emph{a posteriori} category.
There are, however, at last three reasons for taking preference information into account earlier~\citep{Branke2016}:
\begin{enumerate}
\item It will allow to provide the DM with a more relevant sample of Pareto optimal alternatives. This could either be a smaller set of only the most relevant (to the DM) alternatives, or a more fine-grained resolution of the most relevant parts of the Pareto frontier.
\item By focusing the search onto the relevant part of the search space, we expect the optimization algorithm to find these solutions more quickly. This is particularly important in computationally expensive applications such as hyperparameter optimization.
\item As the number of objectives increases, it becomes more and more difficult to create an approximation to the complete Pareto optimal frontier. This is partly because of the increasing number of Pareto optimal solutions, but also because with an increasing number of objectives, almost all solutions in a random sample of solutions become non-dominated~\citep{ishibuchi2008evolutionary}, rendering dominance as selection criterion less useful. DM preference information can re-introduce the necessary order relation.
\end{enumerate}
Several ways to specify preferences have been proposed, including reference points (an ``ideal'' solution), constraints (minimum acceptable qualities for each objective), maximal/minimal trade-offs (how much is the DM willing to sacrifice at most in one criterion to improve another criterion by one unit) and desirability functions (non-linear scaling of each objective to $[0,1]$).
Rather than asking the DM to specify preferences explicitly, preferences can also be learned~\citep{furnkranz2010preference} by asking the DM to rank pairs or small sets of solutions, or to pick the most preferred solution from a set. This has the advantage that the DM only has to compare \emph{solutions}, which is something they should be comfortable doing.
Recent efforts have been made to make the multi-objective optimization process itself more explainable and trustworthy through an interactive preference learning approach.
One example can be found in~\citet{misitano2022towards}, where Shapley values are used to help the DM understand trade-offs between objectives in each iteration, enabling them to articulate preferences in an informed manner.
Finally, people have observed that DMs often select a solution from the Pareto frontier that ``sticks out'' in the sense that improving it slightly in either objective would lead to a significant deterioration in the other. These solutions are often called ``knees'', and it is possible to specifically search for them without having to ask the DM anything (\citep{BDDO04}).

\subsubsection{Noisy environments}
If an MOO problem is noisy, we do not have direct access to the objective values $c_1, \dots, c_m$, but instead we only have measurements $\tilde c_1, \dots, \tilde c_m$ with 
\begin{equation}
    \tilde c_i = c_i + \epsilon_i\quad \forall i = 1,\dots, m 
\end{equation}
where $\epsilon_i$ is the observational noise in the $i$-th objective modeled as a random variable. 
Noise in the evaluation plays a major role in most MOHPO problems, because generalization error estimates in Eq.~(\ref{eq:GEH}) are based on a finite dataset sampled from a much larger universe,  the random sequence of the data as it is presented during training has an impact on the performance, or a stochastic optimizer is used during training.
Clearly, noise is challenging for optimization, whether single-objective or multi-objective. It  may lead to false performance comparisons such as an incorrectly inferred dominance relationship between two HPCs. This may mean some dominated solutions are classified as non-dominated and thus incorrectly returned by the algorithm, while some Pareto-optimal solutions are discarded incorrectly because they are perceived as dominated. It may also lead to an over-optimistic estimate of the Pareto frontier, as most solutions returned as non-dominated  were ``lucky'' in their evaluation. 

One simple way to reduce the effect of noise is to evaluate each solution multiple times and optimize based on mean values. While this reduces the standard error, it is computationally very expensive and may be impractical for HPO.
Researchers in the evolutionary computation community have developed a wealth of methods to cope with noisy evaluations, including the use of statistical tests (e.g.,~\citet{syberfeldt2010noisy},~\citet{park2011noisy}), the use of surrogate models (e.g.,~\citet{BSS01}), probabilistic dominance~\citep{Hughes2001}, or integrating statistical ranking and selection techniques~\citep{hay2009}.
A relatively simple yet effective method seems to be the rolling tide EA~\citep{fieldsend2015RTEA} which alternates between sampling new candidates and refining the archive, i.e., re-evaluating promising HPCs, in each optimization iteration. 
For BO, the noise can be accounted for by using re-interpolation~\citep{koch2015efficient} or in a straightforward way by  choosing GP regression (rather than interpolation) and appropriate acquisition functions, see, e.g.,~\citet{Knowles09,ROJASGONZALEZ2020,ROJASGONZALEZ2020survey,horn2017noisy,astudillo17,hernandez2016predictive,Daulton21}. 
A recent survey on multi-objective optimization methods under noise with a provable convergence to a local non-dominated set has been provided by~\citet{Hunter19}, older surveys with a somewhat broader scope can be found in~\citet{JinBranke05,gutjahr2016stochastic}.

\subsubsection{Realistic evaluation of multi-objective optimization methods}

When applying MOHPO in a real world setting it is crucial to understand how a solution will behave on unseen data after deployment. 
In standard HPO this is achieved via a 3-way split of the data into train, validation and test sets, or, more generally, as nested resampling (c.f. Section~\ref{sec:foundations}).
In Section~\ref{ssec:qual_ind} different measures for the quality of multi-objective optimization have been introduced. 
While these measures are generally useful to evaluate the performance over the whole objective space, decision makers are ultimately only interested in a single solution (selected from the Pareto set) for deployment and later use.
This usually implies a human-in-the-loop. 
For a single train/validation/test split, this is easily achievable: 
The decision maker needs to look at the Pareto front computed on the validation set, choose the configuration they would like to use and then evaluate its performance on the test set. 
Extending this approach to nested resampling, multiple Pareto fronts are generated, one for each outer fold. 
Here, a drill-down to a single solution would imply that the decision maker needs to make these choices for each outer loop, which can become impractical and can make larger benchmark studies difficult to conduct efficiently.
What can be implemented in an automatic fashion, is the evaluation of each generated Pareto front on its associated outer test set in an unbiased fashion.
Here, similarly as in nested resampling in single-objective HPO, each Pareto set candidate would be trained on the joint training and validation set, and evaluated on the test set. This results in a new unbiased Pareto front for each outer iteration, and measures like hypervolume can also be calculated from the outer results in an unbiased fashion.
In general, proper MOHPO evaluation is an understudied and open challenge for further research.


\subsection{Relevant benchmarks and results}
Most benchmarks comparing different multi-objective optimizers for HPO were conducted in the context of new optimizers being proposed (see e.g.,~\citep{schmucker2020multi_obj_multi_fid,hernandez2016predictive,guerrero2021bag_of_baselines}).
To the best of our knowledge no extensive and dedicated benchmark of MOHPO methods covering all (or even a large number of) relevant scenarios of MOHPO have been published.
That being said, recent standardized HPO benchmark suites like \emph{HPOBench}~\citep{eggensperger_hpobench} and \emph{YAHPO Gym}~\citep{pfisterer_yahpo} include support for selected multi-objective use cases, which emphasizes the trend towards MOHPO in the research community and enables further experimentation and research.
Where the former only supports multiple prediction performance objectives, the latter includes a number of objectives related to computational efficiency and interpretability for some scenarios.
Another benchmark, \emph{EvoXBench} has been introduced quite recently and offers insight into the performance of various EAs applied to NAS problems (image classification)~\citep{lu2023neural}.
Table \ref{mohpo_benchmarks} presents a selection of relevant papers that include a MOHPO benchmark or an examination of various algorithms on selected tasks in a meticulous and transparent manner. 
{\renewcommand{\arraystretch}{1.4}
\begin{table}[]
\begin{tabular}{>{\centering\arraybackslash}p{1.5cm}p{5cm}>{\centering\arraybackslash}p{2cm}p{3.5cm}}
\hline
Reference                                               & Algorithms                                                                                                                                         & \# Scenarios & Note                                                                                                             \\ \hline
\citep{pfisterer_yahpo}               & Random Search, ParEGO, SMS-EGO, EHVI, Multi-EGO, Mixed integer evolution strategy                    & 25           &     Surrogate benchmark, focus on mixed spaces with hierarchical structures                                                                                                             \\ \hline
\citep{horn2016multi}                  & ParEGO, SMS-EGO, NSGA-II, Latin Hypercube Sampling                                                    & 9            & Only binary classification                                             \\ \hline
\citep{horn2017noisy}                  & SMS-EGO, Rolling tide EA, Random search                                                                     & 9            & Only SVMs in binary classification and emphasis on noisy scenarios \\ \hline
\citep{hernandez2016predictive}        & PESMO, EHVI, SMS-EGO, ParEGO, Sequential uncertainty reduction                                      & 1            &                                                                                                                  \\ \hline
\citep{guerrero2021bag_of_baselines} & EMOA with successive halving, Multi-objective BOHB, EHVI, Multi-objective BANANAS, BULK \& CUT optimizer & 2            &    \\ \hline            
\citep{lu2023neural}                  & NSGA-II, NSGA-III, MOEA/D, IBEA, HypE, RVEA                                                                     & 18           & Focus on EAs for NAS, benchmark suite supports more instances
                                                                                         \\ 
\hline
\end{tabular}
\caption{Overview over publications that include relevant MOHPO benchmarks and/or meticulous comparisons between different algorithms, that are set up in a transparent manner.}
\label{mohpo_benchmarks}
\end{table}}
While some findings in the presented publications should be taken with a grain of salt (as they may be restricted to very specific scenarios), we try to extract some general lessons and best practices from the current body of literature.

\paragraph{1. Random search is a competitive baseline and preferred over grid search.}
The experiments conducted in~\citet{horn_thesis2019} reveal the weaknesses of utilizing grid search in the context of multi-objective HPO when tuning hyperparameters of an SVM for binary classification (classification error and training time as targets) compared to sequential model-based optimization techniques in the sense that grid search fails to provide a good Pareto front approximation. 
Random search on the contrary has shown good results and even outperformed some model-based optimization methods (i.e., ParEGO, SMS-EGO and PESMO) when applied to a fairness-related multi-objective HPO task~\citep{schmucker2020multi_obj_multi_fid}.
Findings in \citet{pfisterer_yahpo} support this: Multi-EGO, ParEGO and mixed integer evolution strategy are on par with or outperform random search, while EHVI or SMS-EGO sometimes fail to outperform the vanilla random search.

\paragraph{2. Different horses for different courses.}
The optimal choice of optimization algorithm heavily depends on the MOHPO problem at hand.
For example, while in the experiments in \citet{horn2016multi}, ParEGO and SMS-EGO were able to outperform an NSGA-II variant and Latin Hypercube Sampling, the experiments in~\citet{horn2017noisy} showed a rolling tide EA to be quite competitive against SMS-EGO and outperforming random search.
This notion is confirmed in the benchmarks from \citet{pfisterer_yahpo}, as their results indicate strong performance differences of optimizers with respect to the hypervolume indicator.
For example, mixed integer evolution strategy performs exceptionally well on some benchmark problems but only shows average or below par performance on others.
This can also be seen in~\citet{lu2023neural}, where several EAs are tested on a number of NAS instances.

\paragraph{3. Sample efficiency of BO is a big factor.}
Many MOHPO problems, like NAS, are quite expensive to evaluate repeatedly~\citep{lu2023neural}.
As in single-objective HPO, BO methods are appealing because of their sample efficiency. \citet{hernandez2016predictive} examine various BO methods and introduce PESMO for tuning hyperparameters of a feed-forward neural network on the MNIST dataset with classification error and prediction time as objectives. In the context of this expensive task, they found good performance of BO methods - and especially PESMO - on fewer iterations.\\
\\
Even though these benchmarks and especially the benchmark study in \citet{pfisterer_yahpo} are first steps towards an exhaustive MOHPO benchmark and a better understanding of optimizers' behavior on a variety of tasks, plenty of work still needs to be done to be able to give general recommendations regarding when to use which multi-objective optimizer.

\section{Objectives and applications}
\label{sec:applications}

\begin{figure}
    \centering
\begin{tikzpicture}[scale=0.46, every node/.style={transform shape},
    shorten >=0.5pt, >={Stealth[round]},
    hbox/.style = {rectangle, rounded corners = 2, node distance = 2.5cm, fill = white, draw = black, thick, text width = 9em, align = center,
        blur shadow = {shadow blur steps = 5}, minimum height = 1cm, minimum width = 1.5cm, font = \large
},
    sbox/.style = {rectangle, rounded corners = 2, node distance = 1.5cm, fill = white, draw = black, thick, text width = 8em, align = center,
        blur shadow = {shadow blur steps = 5}, minimum height = 1cm, minimum width = 1.5cm
}]
x
\node[hbox, text width = 14em, minimum height=1.5cm] (application) at (0, 0) {\huge Application Domains};
\node[hbox, below of = application, xshift = -12.5cm, text width = 11em, minimum height=1.7cm] (performance) {\huge Prediction performance};
\node[hbox, below of = application, xshift = -7.5cm, text width = 11em, minimum height=1.7cm] (efficiency) {\huge Computational efficiency};
\node[hbox, below of = application, xshift = -2.5cm, text width = 11em, minimum height=1.7cm] (fairness) {\huge Fairness};
\node[hbox, below of = application, xshift = 2.5cm, text width = 11em, minimum height=1.7cm] (interpretability) {\huge Interpretability};
\node[hbox, below of = application, xshift = 7.5cm, text width = 11em, minimum height=1.7cm] (robustness) {\huge Robustness};
\node[hbox, below of = application, xshift = 12.5cm, text width = 11em, minimum height=1.7cm] (sparseness) {\huge Sparseness via feature selection};
\draw[-narrowtriangle, line width = 0.3mm, black] (application.south) -- (0, -1) -- (performance.north |- , -1)  -- (performance.north);
\draw[-narrowtriangle, line width = 0.3mm, black] (application.south) -- (0, -1) -- (efficiency.north |- , -1)  -- (efficiency.north);
\draw[-narrowtriangle, line width = 0.3mm, black] (application.south) -- (0, -1) -- (fairness.north |- , -1)  -- (fairness.north);
\draw[-narrowtriangle, line width = 0.3mm, black] (application.south) -- (0, -1) -- (interpretability.north |- , -1)  -- (interpretability.north);
\draw[-narrowtriangle, line width = 0.3mm, black] (application.south) -- (0, -1) -- (robustness.north |- , -1)  -- (robustness.north);
\draw[-narrowtriangle, line width = 0.3mm, black] (application.south) -- (0, -1) -- (sparseness.north |- , -1)  -- (sparseness.north);

\node[sbox, below right of = performance, yshift = -1.5cm] (roc) {\LARGE Accuracy};
\node[sbox, below right of = performance, yshift = -3.5cm] (nlp) {\LARGE True Positive Rate};
\node[sbox, below right of = performance, yshift = -5.5cm] (objd) {\LARGE Word \\ Error Rate};
\node[sbox, below right of = performance, yshift = -7.5cm] (dots) {\LARGE $\ldots$};
\draw[-narrowtriangle, line width = 0.3mm, black] (performance.text |- performance.south) |- (roc.west);
\draw[-narrowtriangle, line width = 0.3mm, black] (performance.text |- performance.south) |- (nlp.west);
\draw[-narrowtriangle, line width = 0.3mm, black] (performance.text |- performance.south) |- (objd.west);
\draw[-narrowtriangle, line width = 0.3mm, black] (performance.text |- performance.south) |- (dots.west);

\node[sbox, below right of = efficiency, yshift = -1.5cm] (flop){\LARGE Energy\\Consumption};
\node[sbox, below right of = efficiency, yshift = -3.5cm] (model){\LARGE Memory\\Consumption};
\node[sbox, below right of = efficiency, yshift = -5.5cm] (inference){\LARGE Prediction and Training Time};
\draw[-narrowtriangle, line width = 0.3mm, black] (efficiency.text |- efficiency.south) |- (flop.west);
\draw[-narrowtriangle, line width = 0.3mm, black] (efficiency.text |- efficiency.south) |- (model.west);
\draw[-narrowtriangle, line width = 0.3mm, black] (efficiency.text |- efficiency.south) |- (inference.west);

\node[sbox, below right of = fairness, yshift = -1.5cm] (odds) {\LARGE Equalized Odds};
\node[sbox, below right of = fairness, yshift = -3.5cm] (opps) {\LARGE Equality of Opportunity};
\node[sbox, below right of = fairness, yshift = -5.5cm] (cali){\LARGE Calibration};
\draw[-narrowtriangle, line width = 0.3mm, black] (fairness.text |- fairness.south) |- (odds.west);
\draw[-narrowtriangle, line width = 0.3mm, black] (fairness.text |- fairness.south) |- (opps.west);
\draw[-narrowtriangle, line width = 0.3mm, black] (fairness.text |- fairness.south) |- (cali.west);

\node[sbox, below right of = interpretability, yshift = -1.5cm] (spar) {\LARGE Sparseness};
\node[sbox, below right of = interpretability, yshift = -3.5cm] (int) {\LARGE Complexity of Main Effects};
\node[sbox, below right of = interpretability, yshift = -5.5cm] (comp) {\LARGE Interaction Strength};
\draw[-narrowtriangle, line width = 0.3mm, black] (interpretability.text |- interpretability.south) |- (spar.west);
\draw[-narrowtriangle, line width = 0.3mm, black] (interpretability.text |- interpretability.south) |- (int.west);
\draw[-narrowtriangle, line width = 0.3mm, black] (interpretability.text |- interpretability.south) |- (comp.west);

\node[sbox, below right of = robustness, yshift = -1.5cm] (dom) {\LARGE Distribution Shift};
\node[sbox, below right of = robustness, yshift = -3.5cm] (adv) {\LARGE Adversarial Examples};
\node[sbox, below right of = robustness, yshift = -5.5cm] (pert) {\LARGE Perturbations};
\draw[-narrowtriangle, line width = 0.3mm, black] (robustness.text |- robustness.south) |- (dom.west);
\draw[-narrowtriangle, line width = 0.3mm, black] (robustness.text |- robustness.south) |- (adv.west);
\draw[-narrowtriangle, line width = 0.3mm, black] (robustness.text |- robustness.south) |- (pert.west);

\node[sbox, below right of = sparseness, yshift = -1.5cm] (nfeat) {\LARGE Number of Features};
\node[sbox, below right of = sparseness, yshift = -3.5cm] (feats) {\LARGE Stability of Feature Seletion};
\draw[-narrowtriangle, line width = 0.3mm, black] (sparseness.text |- sparseness.south) |- (nfeat.west);
\draw[-narrowtriangle, line width = 0.3mm, black] (sparseness.text |- sparseness.south) |- (feats.west);

\draw[draw = green!60!black, dashed, line width = 1.5, rounded corners = 4] (14.7, -10.4) -- (15.5, -10.4) -- (15.5, -1.3) -- (-4.7, -1.3) -- (-4.7, -10.4) -- (7.8, -10.4);
\node[font = \Large, fill = none] at (11.24, -10.4) [text = green!60!black] {\textbf{FAT-ML related objectives}};
\end{tikzpicture}
    \caption{Overview of application scenarios for MOHPO.}
    \label{fig:taxonomy_applications}
     \Description{Division of application domains into prediction performance, computational efficiency, fairness, interpretability, robustness and sparseness via feature selection. (The last four items are FAT-ML related objectives) Division of predictive performance into ROC analysis, natural language processing and many others. Division of computational efficiency into energy consumption, memory consumption, prediction and training time. Division of fairness into equalized odds and equality of opportunity. Division of interpretability into sparseness, interaction strength, complexity of main effects. Division of Robustness into distribution shift, adversarial examples and perturbation. Division of sparseness via feature selection into number of features and stability of feature selection.}
\end{figure}
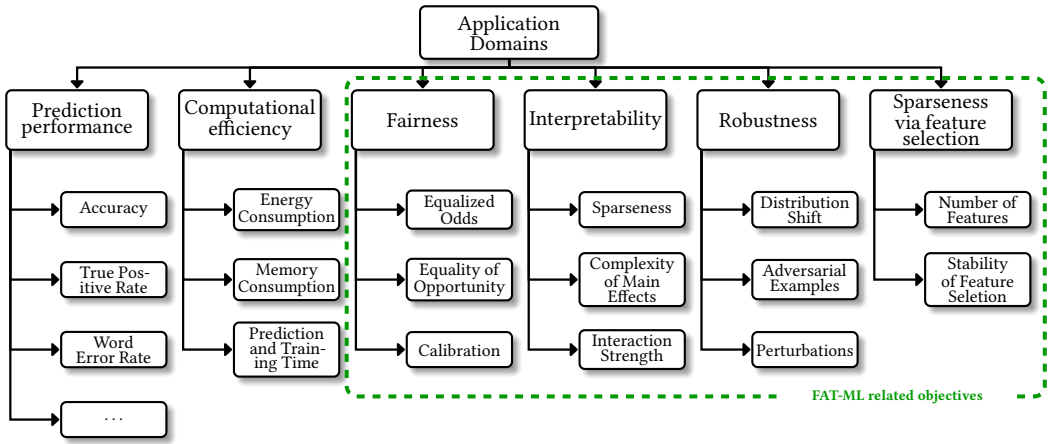

\tikzset{
  basic/.style  = {draw, text width=4cm, font=\sffamily \scriptsize, rectangle},
  root/.style   = {basic, rounded corners=2pt, thin, align=center, fill=white},
  level-2/.style = {basic, rounded corners=3pt, thin,align=center, fill=white, text width=2.3cm},
  level-3/.style = {basic, thin, align=center, fill=white, text width=1.8cm},
  root2/.style = {root},
  level-4/.style = {basic, thin, align=center, fill=white, text width=1.6cm},
  place/.style  = {basic, rounded corners=0pt, align=center, text width=0cm}
}

In the following, we will give an overview of relevant metrics for ML models, interesting use cases and application domains for MOHPO. 
We organize this section by examining three aspects of ML model evaluation: 

\paragraph{Prediction performance}

In most cases, prediction performance is of primary importance. Which performance metric aligns best with the goals and costs associated with an ML task is not always readily apparent, especially if misprediction costs are hard to quantify or even unknown. 
We discuss the case where multiple performance metrics with unquantifiable trade-offs are relevant by the example of ROC analysis in classification. 

\paragraph{Computational efficiency}

Computational efficiency is a prime example for the difficulty of finding an appropriate metric. The desire for a memory or energy efficient model is often difficult to operationalize because it is not as straightforward to measure model efficiency. We will examine various possible ways and present a use case where efficiency and prediction performance are optimized simultaneously.

\paragraph{Fairness, interpretability, robustness and sparseness}

Many applications areas require ML models to fulfill higher standards than only strong predictive performance~\citep{kaur2022trustworthy}. 
Legal requirements and ethical guidelines have been published and the question of how to regulate artificial intelligence with respect to ethics is still an ongoing debate \citep{eu-ethics-guidelines}.
In general, assessing and measuring the trustworthiness of ML models is a challenge \citep{eu-ethics-guidelines, tailor-trustworthy-ai}; we aim to give an overview over approaches relevant to MOHPO.
First, we explore fairness and interpretability as objectives.
We further examine Robustness of ML models to domain shift, perturbations and adversarial attacks as an objective, since robust models inspire a higher degree of trust.
Finally we will venture into the topic of sparse ML models, where the number of features is minimized in some manner.
Sparseness itself is not necessarily a desirable quality, but may be a suitable proxy for model complexity/interpretability and/or data acquisition cost and may even impact performance.
The term FAT-ML (Fairness, Accountability and Transparency in Machine Learning) has been coined and has gained some traction recently\footnote{See e.g., \url{https://facctconference.org/} or \url{https://www.fatml.org/}.}.
There is considerable overlap between the objectives examined here and FAT-ML, but FAT-ML also entails external characteristics, such as e.g., responsibility which is encompassed in the accountability aspect.\\
\\
A visual representation of relevant evaluation criteria and the remainder of this section can be found in Figure~\ref{fig:taxonomy_applications}.
It should be noted that interactions between different objectives are worth studying as well. Objectives may be positively correlated, for example a model with high predictive accuracy might be expected to also perform well measured in other performance metrics like AUC~\citep{caruana2004data}. Similarly, a sparse model is expected to be more efficient as well as more interpretable. Objectives might also be conflicting. A more complex model might for example perform better in terms of accuracy, but might be less interpretable. 
It is hard to quantify these relationships without a comprehensive benchmark, but we will try to shed light on this topic wherever relevant and discuss the open challenges associated with it in Section~\ref{sec:discussion}.

\subsection{Prediction performance}

HPO has traditionally focused on optimizing for predictive performance measured by a single performance metric. 
It can, however, also be beneficial to optimize multiple different prediction metrics simultaneously, in particular if a trade-off between different metrics cannot be specified \emph{a priori}. 
An abundance of prediction metrics have been defined for various ML tasks and the appropriate choice depends on the specific use case and available data and we will therefore not provide a comprehensive overview. 
A good examples for the diversity in metrics can be found for e.g., classification, one of the most widely utilized ML tasks, in Figure 3.1 of~\citet{japkowicz2011classification_evaluation}.
Different performance metrics may penalize prediction errors in distinct ways, and therefore prediction metrics are more or less correlated~\citep{zhou2014correlation_performance_metrics, caruana2004data}.
These correlations have been extensively studied for some (widely used) metric pairs~\citep{huang2005using, cortes2003auc, davis2006relationship}.
We will in the following explore the popular use case of ROC analysis as a typical example where one faces two prediction performance metrics. 
We then briefly introduce two other applications, where this is frequently the case to make the issue more tractable.

\subsubsection{ROC analysis}
\label{ssec:roc}

\begin{figure}
    \centering
    \includegraphics[width=0.7\textwidth]{"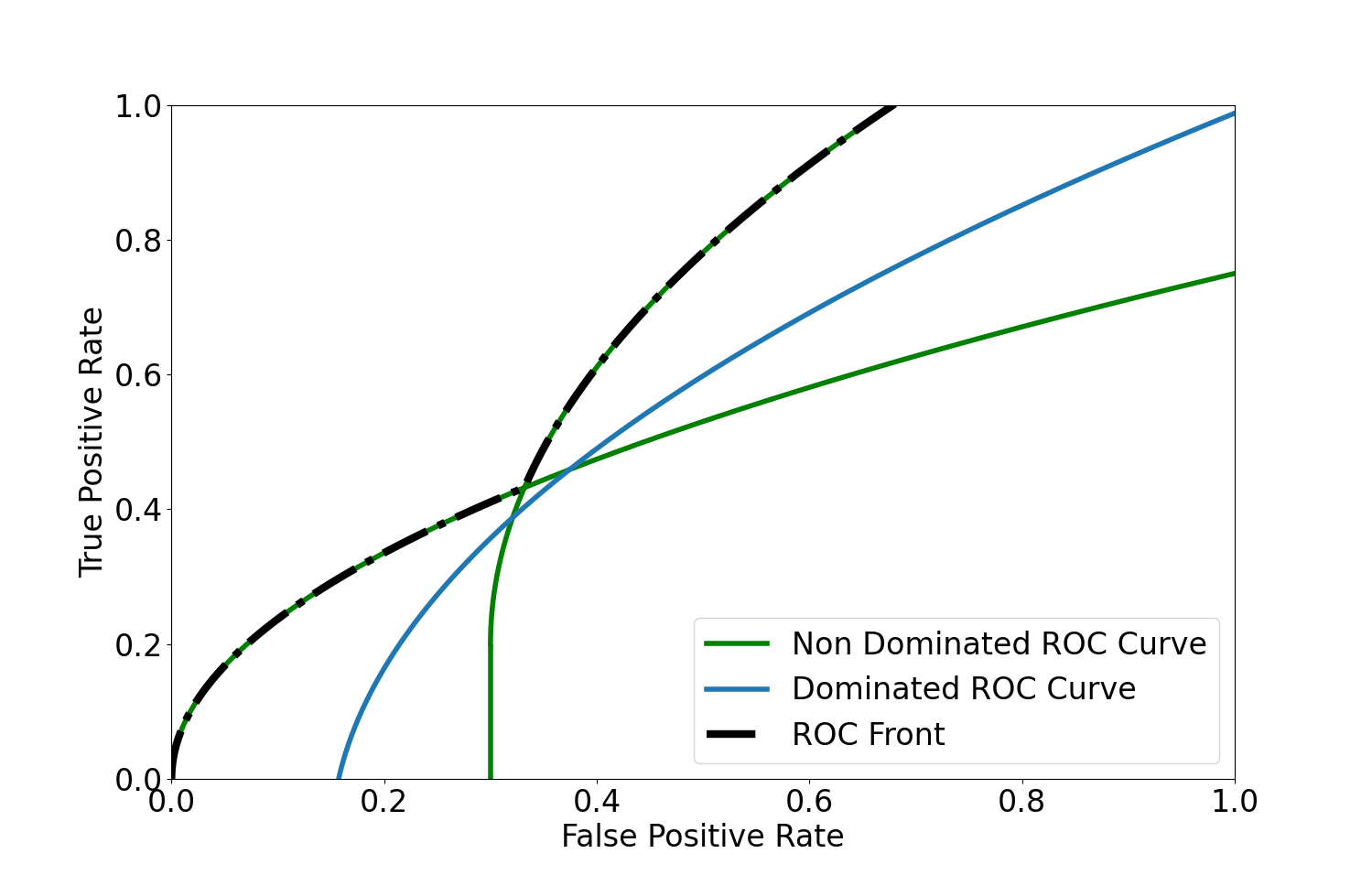"}
    \caption{Scheme showing two intersecting, non-dominant ROC curves with similar AUC in green and one ROC curve that is fully dominated in blue. The resulting Pareto front is marked through a dashed black line.}
    \Description{3 ROC curves are shown. The resulting Pareto front is the the union of the non-dominated parts of each Pareto front.}
    \label{fig:roc-front}
\end{figure}

Many binary classification models predict scores or probabilities that are then converted to predicted classes by applying a decision threshold. 
Different decision threshold values will result in different trade-offs between performance measures. A higher decision threshold may for example reduce the number of false positives, but may also reduce the number of true positives at the same time. 
This trade-off is often visualized by the receiver operating characteristic (ROC) curve, where true positive rate (TPR) is plotted against false positive rate (FPR) for different threshold values\footnote{Similarly, other curves and trade-offs may be analyzed. Another prominent example is the precision-recall curve, which visualizes true positive rate versus positive predicted value.}.
Because improving one classification metric by varying the decision threshold is typically associated with deteriorating performance with respect to the other, choosing a decision threshold based on ROC analysis is fundamentally a multi-objective problem, where the decision threshold can be viewed as an additional hyperparameter. As the decision threshold has no impact on the model itself, it can be optimized post-hoc; tuning it in this manner is fairly cheap~\citep{bischl2021hyperparameter}.
An approach to address this optimization problem is to formulate it as a single-objective problem by e.g., aggregating elements of the confusion matrix (\textit{F-Measure}) or the ROC curve (AUC) into a single metric.
However, not all information is preserved. An example can be seen in Figure~\ref{fig:roc-front}, where two ROC curves lead to a similar AUC but present quite distinct shapes.
It may be desirable to consider the ROC curves for different hyperparameter configurations for a final solution.
One can follow the approaches in e.g.,~\citet{levesque2012multi},~\citet{chatelain2010multi}, and~\citet{bernard2016multiclass} and combine the information from each iteration in one ROC front that displays all Pareto-optimal trade-offs between TPR and FPR. 
This preserves all relevant information from individual evaluations and allows for decisions in accordance to user- or case-specific preferences.
In our example in Figure~\ref{fig:roc-front}, the final combined model would then show the same performance, i.e., trade-offs, as the dashed black line.
A similar approach has been introduced for regression: The Regression Error Characteristic (REC) Curve examines trade-offs between error tolerance and percentage of points within the tolerance~\citep{bi2003regression}.\\
\\
Extending these ideas to a multi-class setting, where each class has unique associated misclassification costs, brings new challenges.
The number of different misclassification errors grows quadratically with the number of classes $g$, leading to increasingly high-dimensional surfaces. Naturally, one wants to minimize all the different misclassification errors simultaneously. 
An in-depth exploration of this challenge and how to solve it can be found in~\citet{everson2006multi}:
The authors define the ROC surface for a $g$-class classification problem and try to optimize for the respective $g(g-1)$ misclassification rates by an EA.
Multi-class ROC problems are therefore generally seen as a MOO problem and closely related to the core topic of this work~\citep{everson2006multi,fieldsend2005formulation}.
They are generally solved through (i) single-model approaches, where one classifier is identified, and once the costs are known at prediction time, a suitable trade-off on that classifier's ROC surface is then found, or (ii) multi-model approaches that produce a pool of suitable classifiers that are available at prediction time~\citep{bernard2016multiclass}.
The optimal ROC surface can be viewed as a Pareto front; in this case the search space is the space of classifiers, and every classifier corresponds to a ROC curve. A classifier is non-dominated if any part of its ROC curve is non-dominated and the length of the non-dominated part of the ROC can be used as the crowding distance.
It should be noted that the principle of AUC as a metric of the classifier's general ability to discriminate both classes in the binary case does not carry over to the multi-class setting~\citep{edwards2005hypervolume}.
\citet{hand2001} attempt to generalize and adapt AUC from a binary setting to a meaningful metric for the multi-class setting.
In binary classification, the ROC curve is generated through varying the decision threshold, which is quite cheap and can be done after each iteration in a hyperparameter optimization problem. 
In a $g$-class setting during prediction, the classifier will generally output a vector of probabilities or some measure of confidence that a sample belongs to the corresponding class $c_i$~\citet{bernard2016multiclass}:
\begin{equation*}
h(x) = [h(c_1|x), h(c_2|x),\dots,h(c_g|x)]
\end{equation*}
To generate a decision rule, a vector of weights $w = (w_1, w_2, ..., w_g)$ is applied, and the highest value is chosen. 
To obtain a ROC surface similar to the ROC curve in binary classification, a large number of different weight vectors would have to be evaluated, which incurs larger additional costs and makes it no longer trivial to obtain post-hoc for each configuration~\citep{bernard2016multiclass}.

\paragraph{Applications and example use case}
An illustrative example is given by~\citet{chatelain2010multi}, who attempt to identify well-performing configurations of SVMs in a binary classification setting with unknown misclassification costs.
Their use-case is based on digit recognition from incoming handwritten mail document images.
While shown to be effective at similar tasks~\citep{neves2011svm}, SVMs are notorious for extreme effects of hyperparameters on performance~\citep{chatelain2010multi}.
\citet{chatelain2010multi} introduce two parameters $C_-$ and $C_+$ as penalties for misclassifying the respective classes (namely, digit vs. non-digit) and use them during training of the SVM. 
They tune for these two hyperparameters along with $\gamma$, the kernel parameter for the radial basis function.
Using NSGA-II (see Section~\ref{sssec:prominent_moeas}), they evolve a pool of non-dominated hyperparameter configurations, thus approximating the Pareto optimal set.
Another application can be found in~\citet{horn2016multi}, where \emph{ParEGO} and \emph{SMS-EGO} (see Section~\ref{ssec:model_based_optimization}) have been applied to jointly minimize the false-negative rate and false-positive rate of an SVM on a variety of binary classification tasks from different domains.


\subsubsection{Natural language processing}
Language and human speech in general are quite complex and in turn natural language processing (NLP) tasks are notoriously hard to evaluate.
One such example is the recent subfield of natural language generation (NLG):
The quality of the resulting texts has to be quantified, but may depend on the use case and different aspects such as semantics, syntax, lexical overlap, fluency, coherency~\citep{kaster2021global}.
\citet{sai2021perturbation} show that across several popular metrics for NLG tasks, no single metric correlates with all desirable aspects.
In a related application of MOHPO,~\citet{schmucker2021sh} use successive halving to optimize perplexity and word error rate for transformer based language models.

\subsubsection{Object detection}
Given an image, object detection is used to determine whether there are instances of a given type and where they are located~\citep{liu2020deep}.
This introduces aspects of both regression (how close is the proposed bounding box of the object compared to the ground truth?) and classification (are all objects identified correctly?) into the task.
Widely used metrics include precision and recall, but this naturally only focuses on one aspect of the task, as one needs to formerly define when something is a true prediction or a false prediction.
\citet{liddle2010multi} propose to specifically examine two aspects of object detection in a multi-objective manner during evaluation: (i) detection rate $DR = \frac{\text{\# correctly located objects}}{\text{\# objects in the image}}$ and (ii) false alarm rate $FAR = \frac{\text{\# falsely reported objects}}{\text{\# objects in the image}}$.
Aside from these prediction performance metrics, detection speed is often crucial, which makes evaluation with a single metric even more challenging~\citep{liu2020deep}.

\subsection{Computational efficiency}
\label{ssec:efficiency}

Technical constraints have always limited ML research and application, but with the increasing prominence of deep learning, efficiency in ML has become an important topic~\citep{sandler2018mobilenetv2, tan2019efficientnet, tan2019mnasnet, wang2020multi}.
This section will not address efficiency of the actual optimization process -- be it model search, hyperparameter optimization or NAS~\citep{baker2017accelerating, liu2018hierarchical, tan2019efficientnet}, which is an active research area in its own right.
In the context of this work, we see efficiency as a desirable quality of an ML model with a given hyperparameter configuration in terms of computational effort needed for training or prediction\footnote{The optimization process can become more efficient if it focuses on efficient candidates during optimization.
An example of this correlation is the expected improvement per second acquisition function~\citep{snoek_practical_2012} in BO, which often prefers configurations that are quick to evaluate.}.
When taking efficiency into account, a common scenario is the existence of resource limitations which have to be respected, e.g., the memory consumption of the model has to be below the available memory to allow for deployment.
In these scenarios it may be more useful to the practitioner to formulate a constrained (single-objective or multi-objective, depending on the remaining objectives) optimization problem.
Another approach to interpret efficiency in the context of an ML model is \emph{Feature Efficiency}, which will be addressed in Section~\ref{ssec:sparseness}.
When looking at hardware implementation, we can roughly differentiate between energy-efficient and memory-efficient models.
In the following, we will introduce several metrics that can be used to measure a model's efficiency.
We will in the following give an introduction to three broad approaches to quantifying efficiency in the context of machine learning models and present one related use case each.
A comprehensive overview of publications applying MOHPO with at least one objective related to efficiency can be found in Appendix~\ref{efficiency_applications_table}\footnote{While some applications of MOHPO such as prediction performance have too many publications to provide a reasonable and comprehensive overview in the scope of this work, others like e.g., interpretability have only recently been introduced and only few related publications exist. Efficiency in MOHPO is established and at the same time still somewhat novel. We therefore believe this comprehensive review -- to the best of our knowledge the first of its kind -- to be a worthwile contribution.}.

\paragraph{Energy consumption and computational complexity}
Limiting computational complexity reduces the number of operations performed in a model and therefore generally leads to rather energy-efficient models.
\citet{lu2019nsga} give an overview of suitable measures for computational complexity of deep learning models: number of active nodes, number of active connections between the nodes, number of parameters and number of floating-point operations (FLOPs).
From experiments, they conclude that FLOPs are the ideal metric and move on to optimize for it in a multi-objective NAS along with accuracy for image classification architectures.
This matches with other research papers; FLOPs have long been used in ML -- especially in deep learning publications over the last decade -- to describe the complexity or even size of a model, for example, in the introductions of ResNet by~\citet{he2016deep} or ShuffleNet by~\citet{zhang2018shufflenet}.
An alternative to FLOPs is to use Multiply-Accumulate (MAC) operations, but the relationship to FLOPs is roughly linear~\citep{hsu2018monas}.
Another approach to measure energy consumption is through the use of an appropriate simulator like \emph{Aladdin} as introduced in~\citet{shao2014aladdin}. 
It is designed to simulate the energy consumption of NNs given the right information (C code describing the operations performed by the NN) and is used for evaluation of architectures~\citep{hernandez2016designing, reagen2017case}.
\subparagraph{Example Application} -
\citet{wang2019evolving} aim to tackle the issue of complex deep learning models for computer vision and the challenges these extremely deep architectures pose for deployment on e.g., edge devices. 
They employ a multi-objective approach to identify models suitable for image classification that are not only highly accurate, but also cope with a minimal amount of FLOPs. 
Convolutional Neural Networks~\citep{lecun1999object} are a popular choice of model for image classification today, and many complex architectures are used for various computer vision tasks. 
One such example is DenseNet~\citep{huang2017densely} which consists of several blocks of dense layers connected via convolutional layers and pooling layers. \citet{wang2019evolving} uses the the hyperparameters of its dense blocks as the search space for the architecture search.
This includes the number of layers and the growth for the dense blocks as well as typical deep learning hyperparameters, such as maximum epochs or learning rate.
From this search space, a population is initiated, and Particle Swarm Optimization~\citep{sierra2005improving} is used to find a Pareto front.
In experiments on the CIFAR-10 dataset, some of the identified models outperform DenseNet-121 and other DenseNet configurations while being less complex with a smaller number of FLOPs.

\paragraph{Model size and memory consumption}
Especially in deep learning models, model size and efficiency will often go hand in hand, as employing more parameters (i.e., greater model size) generally results in more FLOPs.
The parameters are mostly weights, and their number can be straightforwardly derived in deep learning architectures.
Several publications have used the number of parameters as proxy for the efficiency of a deep learning model.
For example~\citet{howard2017mobilenets} introduce MobileNets, which are specifically designed for efficient deployment on edge devices.
They introduce separable convolutions to reduce the number of parameters needed for a top-performing Convolutional Neural Network.
\subparagraph{Example Application}-
\citet{loni2020deepmaker} present their framework \emph{DeepMaker} to identify efficient and accurate models for embedded devices. 
The objectives are classification accuracy and model size (number of trainable network weights); they also discovered a high correlation between the latter and prediction time.
NSGA-II is used to search the space of discrete hyperparameters (activation function, number of condense blocks, number of convolution layers per block, learning rate, kernel size and optimizer).
Their approach has some similarity to multi-fidelity optimization, as during optimization sampled architectures are only trained for 16 epochs to decide for the optimal choice, but the final performance is reported after training the selected architecture for 300 epochs.
The method is tested on MNIST; CIFAR-10 and CIFAR-100.

\paragraph{Prediction and training time}
The model is usually trained before deployment, or the training is not as time-critical, so we are mostly interested in minimizing prediction time. However, some applications and deployment strategies require frequent retraining of the model. In which case, training time can be a crucial factor as well.
While often a crucial metric, prediction time is very hard to measure reliably due to various differences in the computing environment~\citep{lu2019nsga}.
Prediction time may also correlate strongly with energy efficiency metrics,
\citet{rajagopal2020pso_efficiency} use FLOPs as a proxy for inference latency.
\subparagraph{Example Application}-
PESMO is introduced in~\citet{hernandez2016predictive} and applied to several MOO problems, one of which is finding fast and accurate NNs for image classification on MNIST. 
They tune a variety of hyperparameters: The number of hidden units per layer (between 50 and 300), the number of layers (between 1 and 3),
the learning rate, the amount of dropout, and the level of $\ell_1$ and $\ell_2$ regularization~\citep{hernandez2016predictive}.
The objectives are prediction error and prediction time - measured here as the time it takes for the network to predict 10.000 images. 
A ratio is then computed with the time the fastest network in the search space requires for this task.
PESMO is  compared against SMS-EGO, ParEGO and EHI among others and shows superior performance in terms of hypervolume. 
Their follow-up paper~\citep{hernandez2016designing} deals with a similar use case.\\
\\
\citet{dong2018ppp-net} examine bi-objective NAS for image classification optimizing for classification accuracy and FLOPs, the number of parameters and prediction time respectively.
It may sometimes be relevant to optimize a model with two efficiency metrics along with a primary performance metric; examples for this can be found in~\citet{lu2020muxconv, elsken2018efficient, chu2020reinforced_mohpo}.
\citet{chu2020reinforced_mohpo} present a framework that allows incorporation of constraints with respect to the three objectives in a task of the super-resolution domain: peak signal-to-noise-ratio or structural similarity index as performance metrics, FLOPs, and the number of model parameters.
Note, that this section already delves heavily into the topic of HW-NAS, but only from a multi-objective point of view.
Single-objective and constrained approaches~\citep{ijcai2021p592} are also widespread in addressing these challenges.
\citet{ijcai2021p592} provide a comprehensive survey on HW-NAS for a reader interested in this subtopic; even more detailed in their arXiv version \footnote{\textit{A Comprehensive Survey on Hardware-Aware Neural Architecture Search} available at \url{https://arxiv.org/abs/2101.09336}}.


In context of the full software/hardware stack,~\citet{lokhmotov2018multi} tunes the hyperparameters of MobileNets for test accuracy and prediction time on an image classification task.


\subsection{Fairness}
\label{ssec:fairness}
When algorithmic decisions made by ML models impact the lives of humans, it is important to avoid introducing bias that adversely affects sub-groups of the population~\citep{holstein2019fairness, fairmlbook, kaur2022trustworthy}. 
In practice, a user may want their Machine Learning model to be performant and adhere to ethical or legal requirements to ensure fairness; like e.g., a bank deciding on the acceptance of loan applications.
For a survey on different fairness perspectives and a good overview of related metrics, we refer to two excellent surveys by~\citet{mehrabi2021survey} and~\citet{pessach2020algorithmic}, as well as a benchmark by~\citet{sorelle2019fairness_benchmark} for a more comprehensive take on fair ML than is possible in the scope of this paper. 
The choice of an appropriate metric to mitigate biases and measure fairness of a given model is complex and further depends on the context a decision is made in~\citep{vzliobaite2017measuring}.
While a variety of \emph{causal} and \emph{individual} fairness notions exist, so-called (statistical) group fairness metrics are widely used in practice, since they are easy to implement and do not require access to the underlying (causal) data generating mechanism ~\citep{pessach2020algorithmic}.
Given a protected attribute $A$ (e.g., race), an outcome $X$, and a binary predictor $Y_b$, several criteria can be derived~\citep{fairmlbook}, and we provide three widely used examples:
\begin{itemize} 
 \item \textbf{Equalized Odds (Independence)}:
    \[
    Pr(Y_b = 1 | A = 1, Y = y) = Pr(Y_b = 1 | A = 0, Y = y), y \in \{0, 1\}
    \]
    Fulfilling this essentially requires equal true positive and false positive rates between sub-populations. More generally, independence on the conditional $Y$ is required. \emph{Equalized odds} for a binary predictor is the most relevant example of such an independence measure~\citep{hardt2016equality}.
 \item \textbf{Equality of Opportunity (Sufficiency)}:
     \[
    Pr(Y_b = 1 | A = 1, Y = 1) = Pr(Y_b = 1 | A = 0, Y = 1)
    \]
     This criterion is a relaxation of equalized odds, as only independence on the event $Y=1$, which denotes the advantageous outcome (e.g. getting approved for a loan) here, is required. It therefore constitutes \emph{equality of opportunity}~\citep{hardt2016equality}.
 \item \textbf{Calibration} is another desirable criterion for classifiers, especially in the context of fairness, where calibrated probabilities in all groups may be required. 
 For a classifier $h(x)$ that yields predicted probability, \textit{calibration} requires that:
  \[
  \underset{a \in \{0,1\}}{\forall} \underset{p \in [0;1]}{\forall}  Pr(Y_b = 1| A = a, h(x) = p) = p.
  \]
\end{itemize}


\paragraph{Applications and example use case}

\begin{figure}[ht]
    \centering
    \includegraphics[width=0.65\textwidth]{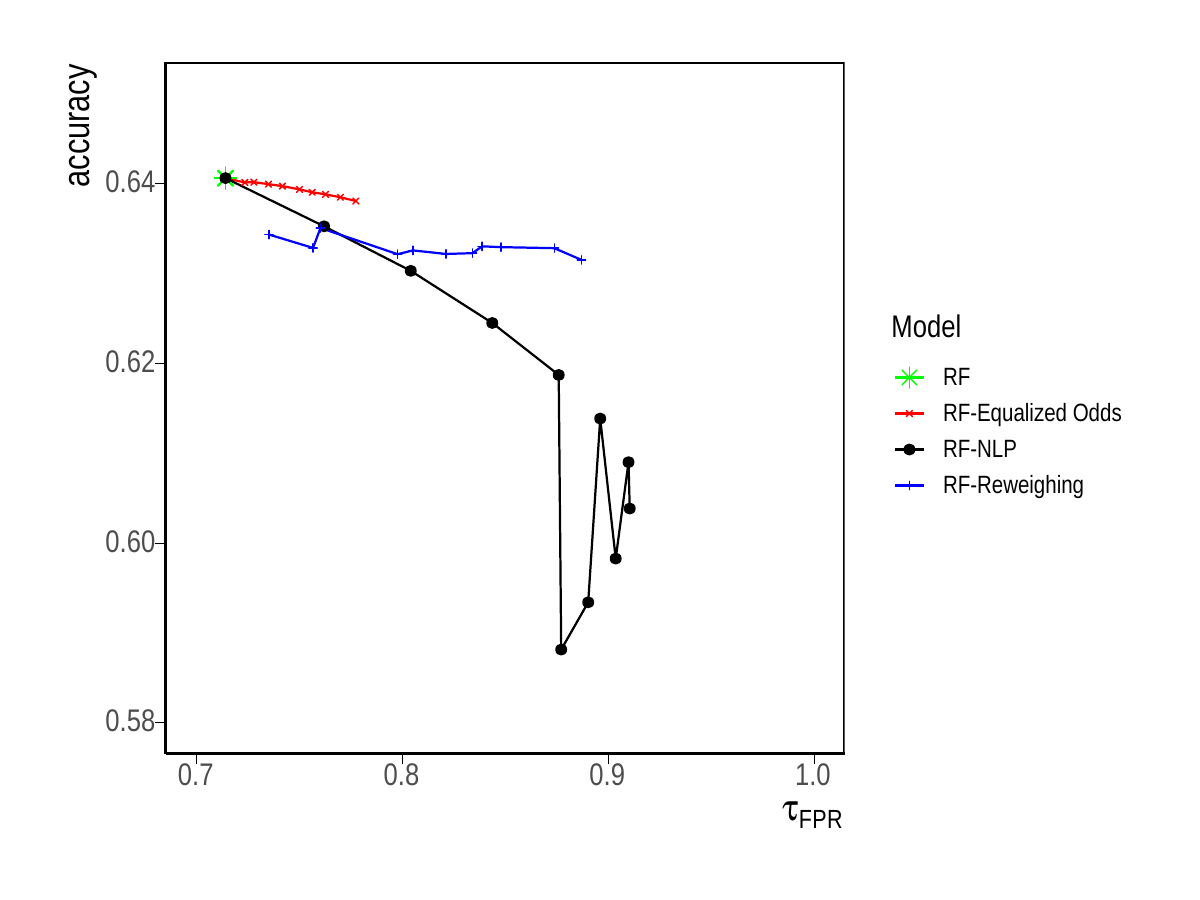}
    \caption{Effect of $3$ different debiasing strategies on the fairness-accuracy trade-off on the COMPAS dataset measured on test data. Figure obtained from~\cite{agrawal21_2}}
    \label{fig:fairness_compas}
    \Description{Three curves are plotted in an accuracy-versus-fairness plot. The performance and fairness value of a standard random forest model are shown in the upper left corner. The curve associated with the reweighting debiasing strategy decreases relatively strongly, while the equalized odds and NLP curves decrease slightly. While the reweighting and equalized odd curve start at the performance and accuracy level of the standard random forest model, the NLP starts with a slightly lower accuracy and slightly higher fairness value. The decreasing order of the highest possible fairness values is: NLP, reweighting, equalized odds.}
\end{figure}

Existing approaches towards improving a ML-algorithm's fairness mainly focus on decreasing discrepancies in classifier performance between sub-groups or individuals (e.g., in~\cite{fairmlbook, lahoti2020fairness}). This is achieved e.g., by preprocessing the data (e.g.,~\cite{kamiran12}), imposing fairness constraints during model fits~\citep{lahoti2020fairness} or by post-processing model predictions~\citep{hardt2016equality, agrawal21_2}.
Those methods in turn often come with hyperparameters that can be further tuned in order to emphasize fairness during training~\citep{schmucker2020multi_obj_multi_fid,schmucker2021sh}.
Several approaches towards optimizing (model-) hyperparameters for fairness in a multi-objective fashion have been proposed. 
While~\citet{pfisterer2019multiobjective} propose multi-objective BO to jointly optimize fairness criteria as well as prediction accuracy,~\cite{Perrone2020} propose a \textit{constrained} BO approach, where the fairness metric is constrained to a small deviation from optimal fairness while predictive accuracy is optimized. 
\citet{pelegrina2020novel} use a MOEA to optimize simultaneously for fairness metrics and performance in an attempt to find a fair PCA.
Yet another approach is introduced in~\citet{martinez2020fairness_moo}, where each sensitive group risk is a separate objective, also leading to a MOHPO problem for choosing a classifier.
It is interesting to note that fairness can be heavily influenced not only by parameters of the debiasing method, but also by the choice of ML algorithm and hyperparameters~\citep{pfisterer2019multiobjective}. 
One popular dataset studied in the context of fairness is the COMPAS (Correctional Offender Management Profiling for Alternative Sanctions) data~\citep{compas}. The goal of COMPAS  is to predict the risk that a criminal defendant will re-offend. The goal is to obtain a prediction that is accurate but simultaneously not biased towards individuals of any race. 
In the following example, the latter quantity is measured via $\tau_{FPR} = \frac{FPR_{S_0}}{FPR_{S_1}}$ (optimal for $\tau_{FPR} = 1$, where $FPR_S$ is the false positive rate on group $S$ with $S_0$ and $S_1$ is the advantaged and disadvantaged group, respectively).
The effect of applying several debiasing strategies -- interpolating between no debiasing and full debiasing -- is shown in Figure~\ref{fig:fairness_compas}. 
A random forest ($RF$) model is trained with different debiasing techniques: Reweighing~\citep{kamiran12}, Equalized Odds~\citep{hardt2016equality}, and nonlinear programs ~\citep{agrawal21_2}. As can be observed, these different strategies and debiasing strengths lead to different trade-offs, thus resulting in a MOO problem. 

\subsection{Interpretability}
\label{ssec:interpretability}

In order to make an ML model's decisions more transparent, different methods that aim at providing human-understandable explanations have been proposed (c.f.~\cite{molnar2019}). 
Requirements for interpretability generally range from models derived from fully interpretable model classes to interpretability via post-hoc interpretation techniques. 
For example, the former is required to satisfy regulatory constraints in the banking sector~\citep{carvalho2019machine, bucker2020transparency} and can be thought of as a constraint on the search space when selecting a model. 
The latter is often used to understand functional relationships between features and target variables in an ML context (i.e., understanding the model) or to understand single decisions made by a trained algorithm~\citep{molnar2019}. 
Interpretability methods can however produce misleading results if a model is too complex or the explainability technique is unreliable due to them e.g., using additional features~\citep{kindermans2019reliability, molnar2019}. 
Quantifying interpretability, i.e., \textit{determining how complex the predictive decisions of a given model are}, is not straightforward, as terms like interpretability, explainability, and complexity are highly subjective expressions~\citep{lipton2018mythos_interpretability, molnar2019, molnar2019quantifying_2, kaur2022trustworthy, tailor-trustworthy-ai}.
A popular, easily quantifiable alternative and often suitable proxy for model complexity (and therefore interpretability) is \emph{sparseness}, i.e., the number of features. 
We explore this concept further in the context of MOHPO in Section~\ref{ssec:sparseness}.
Current approaches to quantifying interpretability focus mainly on tabular data~\citep{molnar2019quantifying_2}.
Therefore, we here present two metrics to quantify interpretability that have been used in practice, originally proposed in~\citep{molnar2019quantifying_2}\footnote{A detailed derivation and explanation can be found in~\citet{molnar2019} and~\citet{molnar2019quantifying_2}}:

\begin{itemize} 
 \item \textbf{Complexity of main effects} \citet{molnar2019quantifying_2} propose to determine the average shape complexity of Accumulated Local Effects (ALE)~\citep{apley2020visualizing} main effects by the number of parameters needed to approximate the curve with linear segments. The Main Effect Complexity (MEC) is defined as:
 \begin{equation}
     MEC = \frac{1}{\sum_{j=1}^p V_j} \sum_{j=1}^p V_j \cdot MEC_j ,
 \end{equation}
 where $V_j = \frac{1}{n} \sum_{i=1}^n (f_{j, ALE}(x^{(i)}))^2$ and $MEC_j = K + \sum_{k=1}^K \mathbb{I}_{\beta_{1,k} > 0} - 1$.
 $K$ is the number of linear segments needed for a good-enough approximation, $p$ is the number of features, $\beta_1, k$ the slope of the $k-th$ linear segment, and $n$ the number of samples.
 \vspace{.1cm}
 \item \textbf{Interaction Strength} Quantifying the impact of interaction effects is relevant when explanations are required, as most interpretability techniques use linear relationships to obtain explanations. Interaction strength $IAS$ can be measured as the fraction of variance that cannot be explained by main effects:
 \begin{equation}
     IAS = \frac{\mathbb{E}(L(f, f_{ALE1st}))}{\mathbb{E}(L(f, f_0))} \geq 0,
 \end{equation}
 where $f$ is the prediction function, $f_{ALE1st}$ the sum of first order ALE effects, and $f_0$ the mean of predictions.
\end{itemize}


\paragraph{Applications and example use case}
\citet{molnar2019quantifying_2} aim to find an accurate and interpretable model to predict the quality of white wines (on a scale from 0 to 10)~\citep{cortez2009wines}.
In order to accomplish this, they define a large search space of several models (SVM, gradient boosted trees, and random forest, among others) with a number of tunable hyperparameters and optimize for four objectives: cross-validated mean absolute error, number of features used, main effect complexity, and interaction strength.
They conduct the optimization with ParEGO (see Section~\ref{sssec:mo-mbo}) over 500 iterations to find good trade-offs between these objectives.
\citet{carmichael2021learning_2} perform MOHPO for deep learning architectures to find optimal trade-offs between accuracy and introspectability for image classification tasks on ImageNet-16-120, CIFAR-10 and MNIST.

\subsection{Robustness}
\label{ssec:robustness}

Despite often hailed as an important requirement for accountability in ML, robustness in the context of an ML algorithm is only loosely defined~\citep{xu2012robustness,cooper2022accountability} and in the case of (MO)HPO sometimes even mixed with the notion of robustness of the optimization procedure~(\citep{kalyan2006robustness}).
For ML algorithms, we broadly distinguish between robustness of the \textit{training procedure} and robustness of the \textit{fitted model}, which have both been described in literature. 
We focus on the latter, i.e., the susceptibility of a \textit{trained model} to shifts of the data in the prediction step. 
While most research into robustness focuses on images, we aim to look at a general case. 
We mostly consider a \textit{classification setting} in the following sections, but we also aim to provide information as to how this differs for \textit{regression} where appropriate.

\subsubsection{Robustness metrics and approaches}

While generally seen as important and relevant, there are no tried and proven metrics to assess the robustness of ML models, nor a proper taxonomy. The taxonomy by~\citet{taori2020measuring} provides an overview of possible changes to the input data. 
We differentiate between three types of changes:

\paragraph{Distribution shift} 
Our notion of a distribution shift refers to changes of either the marginal distribution of the target or the distribution of the features (conditional on the target) on a macro level\footnote{This concept is often associated with domain adaption (which is becoming increasingly popular as a research area, see~\citet{Zhang13} for an introduction)}.
A typical metric used to assess robustness in the context of distribution shifts is the \emph{effective robustness} $\rho(f) = acc_2(f) - \beta(acc_1(f))$, where $acc_1, acc_2$ are the accuracies pre- and post-distribution shift respectively and $\beta(x)$ is the chosen baseline accuracy given a pre-domain shift accuracy $x$.
It represents an adjustment according to the information how much of a performance drop can be expected for models without robustness intervention given a certain accuracy pre-domain shift. 
Additional information on how to compute $\beta(x)$ for a given distribution shift can be found e.g., in~\citet{taori2020measuring}.

\paragraph{Adversarial examples} Adversarial examples have recently generated substantial interest from the visual deep learning community~\citep{goodfellow2015adversarialattacks,akhtar2018adversarial,tramer2017ensemble}, because a model that is very susceptible to adversarial attacks (the use of adversarial examples) is not as trustworthy. 
Adversarial examples are well-known in image data~\citep{wiyatno2019adversarial_images} but have been shown in other types of data, such as text~\citep{huq2020adversarial_text}, sequence~\citep{cheng2020adversarial_sequence}, or tabular data~\citep{ballet2019adversarial_tabular}.
We present two additional metrics initially proposed in~\citet{bastani2016neuralnetrobustness}: 
\begin{itemize} 
 \item \textbf{Adversarial Accuracy} is often used to assess robustness~\citep{tsipras2019robustness, madry2018adversarial_accuracy,mao2019metric}. Adversarial accuracy measures the percentage of samples that are (still) correctly classified after the adversarial attack~\citep{tsipras2019robustness}; for perturbations in an $\varepsilon$-ball around each point, a typical adversarial attack in classification, it can be defined as:
 \begin{equation}
    AAcc = \EX [\vmathbb{1}(f(x^*) = c_x)], \; where \: x^* = \argmax_{d(x',x) \leq \epsilon} L(x', c_x),
 \end{equation}
 where $c_x$ is the respective class label.
 In practice the user would conduct the adversarial attack and then calculate adversarial accuracy from the new predictions.
 \item \textbf{Adversarial Frequency} (originally proposed in ~\citet{bastani2016neuralnetrobustness}) is measured as the accuracy on a worst-case input in an $\ell_{p}$ $\varepsilon$-ball around each point $x_{*}$:
 \[
 AF = P(\rho(f, x_{*}) \leq \varepsilon),
 \]
 where $\rho(f, x_{*})$ is the minimum distance $\hat{\varepsilon}$ for some well-defined metric $d$, so that $\exists \ x,\ d(x,x_{*}) \leq \hat{\varepsilon}: f(x) \neq f(x_{*})$ with $f$ as a classifier.
 \item \textbf{Adversarial Severity} (originally proposed in ~\citet{bastani2016neuralnetrobustness}) is defined as the expected minimum distance to an adversarial example from the input for some $\varepsilon$~\citep{bastani2016neuralnetrobustness}:
 \[
 AS = E[\rho(f, x_{*}) \mid \rho(f, x_{*}) \leq \varepsilon],
 \]
 with $\rho, f, x_{*}$ as before. 
\end{itemize}

While~\citet{bastani2016neuralnetrobustness} deem \textit{adversarial frequency} to be generally the more important metric of the two, significant work centers around the minimum distance to creating an adversarial example, especially for neural networks~\citep{peck2017lowerbounds}. 

\paragraph{Perturbations} Perturbations are often strongly linked to the construction of adversarial examples in deep learning~\citep{laugros2019adversarial}.
This can be seen as a logical connection, as one would often look for an adversarial example within some $\epsilon$-ball.
\citet{laugros2019adversarial} however argue and show that robustness in the context of common perturbations and adversarial robustness are independent attributes of a model. 
One of the most common ways to perturb data is the addition of Gaussian noise, which is also included in their work. 
A simple measure for robustness regarding such a perturbation is presented in~\citet{pfisterer2019multiobjective}. 
Adding random Gaussian noise $N(0, \epsilon)$ \footnote{$\epsilon$ is generally $0.001 - 0.01$ times the range of the numerical feature} to input data $X$ is a typical way to introduce perturbations. One can then compare the loss derived from a relevant measure $L$ on the changed input data to the loss on unperturbed data:
 \[
 |L(X,Y) - L(X+N(0, \epsilon),Y)|.
 \]
Of course, the same procedure can be applied to other types of perturbations. 
In~\citet{pmlr-gilmer2019adversarial}, the authors argue and provide evidence that susceptibility to adversarial examples and perturbations -- at least for image data -- are actually two symptoms of the same underlying problem, and thus, optimizing for robustness against adversarial attacks and more general corruption through perturbations should go hand-in-hand. 
\citet{niu2020evaluating} give a comprehensive overview on perturbations to data in machine translation (mainly synthetic misspellings and letter case changing) and define a variety of suitable measures for model robustness in the context of those perturbations. 

\subsubsection{A note on robustness and uncertainty quantification}
\label{sec:rob_uq}
Uncertainty quantification, especially in the context of deep learning, has become a heavily researched topic and is often mentioned in conjunction with robustness of ML models~\citep{lakshminarayanan2017advances}.
To the best of our knowledge, optimizing an ML model's configuration for proper uncertainty quantification or minimal uncertainty is an under-explored research direction with few exceptions, e.g., \citep{Zaidi_nes} 
-- integrating a respective metric into MOHPO even less so. 
We will therefore restrict this brief section to exploring the connection between uncertainty quantification and robustness as well as other relevant objectives.
Models that are better calibrated with respect to uncertainty tend to suffer less from adversarial examples~\citep{smith2018understanding,lakshminarayanan2017advances}.
Along the same lines, robustness regarding domain shift and predictive performance on out-of-distribution samples are closely linked to (and even used as a measure of) predictive uncertainty~\citep{lakshminarayanan2017advances}.
~\citet{kendall2017uncertainties} show that using the quantification of aleatoric uncertainty has a beneficial effect on the robustness in the context of noise, i.e., perturbations.

\subsubsection{Application and example use case}
The main goal of~\citet{guo2020meets} is to identify network architectures that are robust with respect to adversarial attacks.
To this end, the authors employ one-shot NAS~\citep{bender2018understanding}. 
\begin{figure}[ht]
    \centering
    \includegraphics[width = 0.6\textwidth]{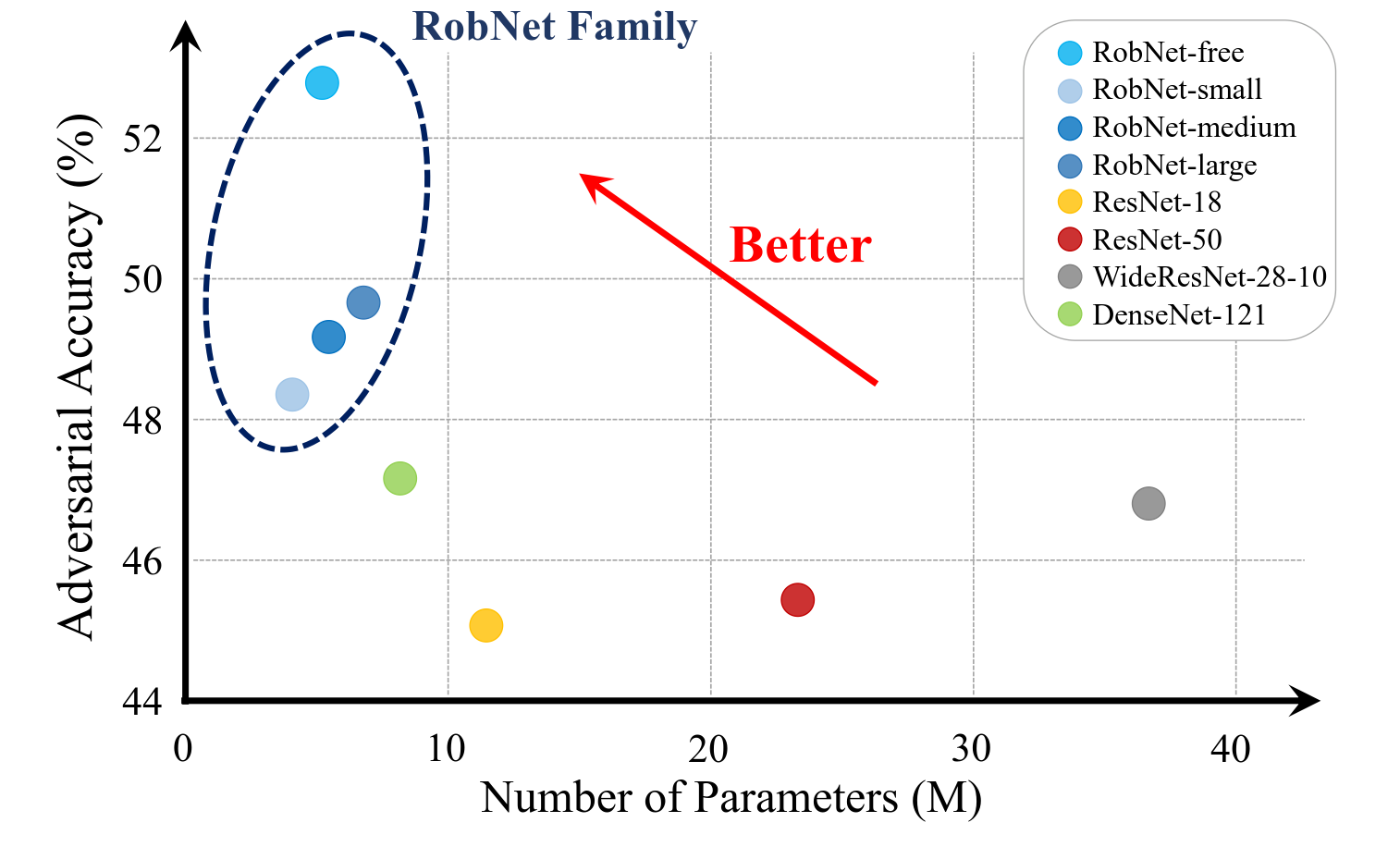}
    \caption{Overview and classification of several architectures examined by~\citet{guo2020meets} with respect to objectives \emph{number of parameters} and \emph{adversarial accuracy}.}
    \label{fig:robnets}
    \Description{An adversarial accuracy versus the number of parameters scatterplot is shown. The whole RobNet family achieves a higher adversarial accuracy while having a smaller number of parameters compared to ResNet-18, ResNet-50, WideResNet-28-10 and DenseNet-121.}
\end{figure}
After initial training of a \emph{supernet}, architectures are drawn through random search, fine-tuned and evaluated. 
The authors then examine different architectures and compare the desirable qualities of robustness in the context of adversarial examples and model size to identify suitable models. This is illustrated for a few select architectures in Figure~\ref{fig:robnets}.

\subsection{Sparseness via Feature Selection}
\label{ssec:sparseness}

In ML, we often face high-dimensional tasks with a large number of features. 
\emph{Sparse} models, i.e., models which use relatively few features, are preferred, because if too many features are considered by a model, (1) the relationship between features and model prediction may be hard to interpret~\citep{Guyon2003},
(2) the cost for model fitting or inference may increase, either in terms of storage, computation, or in terms of actual costs for measuring or acquiring the data~\citep{min2014feature},
(3) the predictive performance of a model might suffer (\emph{curse of dimensionality})~\citep{bellman2015}.
Feature selection is inherently an MOO problem, as model performance and sparsity tend to be conflicting objectives; a lower number of features often means a lower performance due to reduced information.
However, for a certain model and configuration there will be a specific desirable quantity of features, that is probably not the maximum number of features~\citep{Kohavi1997}: 
In the following, we will provide an overview of common feature selection techniques with an emphasis on their connection and possible combination with MOHPO. They are generally divided into three categories~\citep{altashi2020featureselreview}: 

\paragraph{Embedded methods} methods perform feature selection as part of the model fitting process.  
For example, empirical risk minimization with L0 or L1 regularization can shrink irrelevant coefficients to zero, and can therefore automatically produce sparse models during training. 
As another example, trees and tree-based methods can inherently produce sparse models by their greedy split selection.
The drawback of embedded methods is that they are specific to the model in use.

\paragraph{Filter methods} approaches use proxies to rank feature subsets by their likely explanatory power independent of the learning algorithm and before a model is trained. 
Single measures -- for instance, information theoretic measures -- correlation measures, distance measures, or consistency measures ~\citep{Dash1997} are calculated and used for evaluation of different feature subsets.

\paragraph{Wrapper methods} search over the space of selected features to optimize model performance~\citep{Kohavi1997}. 
Because they take the learning algorithm performance into account directly, they often yield better performance than filter methods. 
However, they are computationally more expensive because model performance evaluations are generally noisy and relatively expensive.
Because wrapper methods can use general optimization algorithms, they are the most amenable to extensions to MOO~\citep{altashi2020featureselreview}.

\subsubsection{Sparseness Metrics}
We present the metrics most commonly used in the context of sparseness:
\begin{itemize}
    \item \textbf{(Weighted) Number of Features} included in the model
    $$
        \sum_{j = 1}^p w_j s_j.
    $$
    Minimizing the number of features is the most simple approach, but different weights can be introduced, if features have different costs, e.g., because they have different acquisition costs if a model is intended to be applied in real life.
    \item \textbf{Stability of Feature Selection}: In some applications like e.g., omics data analysis in bioinformatics, the main goal of the analysis is the identification of important features~\citep{LeclercqFeatureOmics2019}.
    The aim is to identify important genes for later examination in the laboratory. 
    Stability of feature selection can be helpful in such cases and is e.g., measured by comparing sets of selected features resulting from different resampling iterations. 
\end{itemize}

In general, the task of feature selection implies search over a binary space $\{0, 1\}^p$ with $s_i = \mathbb{I}[\text{feature } i \text{ is included}]$. To simultaneously perform hyperparameter optimization, we formulate the joint search space of feature and hyperparameter configurations

\begin{eqnarray}
    \{0, 1\}^p \times \LamS. 
    \label{eq:featsel_joint}
\end{eqnarray}

The dimensionality of this space grows exponentially with the number of available features, which makes the optimization problem particularly challenging. 
Additionally, there may be complex interactions between features and hyperparameter configurations. 
Expensive evaluations like wrapper evaluation techniques raise the need for efficient search methods.
Evolutionary algorithms are widely used for feature selection due to their ability to handle complex search spaces,~\citep{abd2014review, xue2015survey}, but recent works have presented BO methods (see Section as a more efficient alternative~\citep{Bommert2020, binder2020mosmafs}.

\subsubsection{Applications and example use case}

\citet{Bommert2017} state feature selection as a MOHPO problem with three objectives: predictive performance, number of features selected, and stability of feature selection. Furthermore, they provide a comprehensive comparison of stability measures.
They tune hyperparameters of the underlying ML pipelines (feature filter plus classification learner) that are relevant to the sparseness objectives via random search to identify desirable trade-offs.
All classification pipelines have been tuned w.r.t.\ the aforementioned three criteria.
In the next step, all configurations that are not within a 5\% tolerance of the best predictive performance identified so far are discarded. 
In comparison with a single-criterion (predictive performance) tuning approach, the multi-objective approach additionally reveals hyperparameter settings that yield models with comparable or even better performance.
At the same time, these additional identified model configurations require fewer features, and the feature selection is more stable \citep{Bommert2017}.
A comprehensive review with additional examples can be found in~\citet{altashi2020featureselreview} and specifically for Evolutionary Computation methods in~\citet{xue2015survey}\footnote{This work mainly includes works that reduce feature selection to a single-objective problem, but also several multi-objective approaches.}.

\subsection{MOHPO applications in industry}
\label{ssec:mohpo_in_industry}
As with single-objective HPO and AutoML, it is hard to gauge the deployment of MOHPO methods in the industry: Companies don't necessarily advertise or quantify their success with such tools. 
This leaves examining the (commercial) tools offered along with relevant publications.
While several libraries like \textit{Syne Tune} or \textit{Optuna} support multi-objective optimization (see also Section~\ref{sssec:bo_software}), commercial AutoML tools have not widely adopted such methods\footnote{Commercial tools tend to focus more on AutoML than pure HPO.}.
The applicability to industrial problems also depends on the metrics involved.
The ML community is still struggling with taxonomies and consequently quantifying notions of, e.g., interpretability, robustness, or fairness~\citep{arrieta2020xai, linardatos2021xai}.
Without established metrics, the productive application of MOHPO techniques is limited. 
On the other hand, the need to jointly optimize pipelines and architectures for performance and efficiency is more established within the general topic of HW-NAS, with several publications emphasizing the importance for industrial applications~\citep{smithson2016neural, schmucker2020multi_obj_multi_fid, ijcai2021p592}.
This intuitively makes sense as embedded devices, and generally, devices that cannot support high compute or energy consumption, are extremely relevant to a number of applications today~\citep{rajagopal2020pso_efficiency}.
When scouring MOHPO-related publications, it can be seen that some companies are involved in MOHPO research, with, e.g., \textit{Amazon} introducing new methods and algorithms in the recent past~\citep{schmucker2020multi_obj_multi_fid, schmucker2021sh}.
However, these methods are not introduced on datasets related to \textit{Amazon's} or other industrial applications, but mainly on typical benchmark datasets~\citep{schmucker2020multi_obj_multi_fid}.
The diversity in applications, especially for performance and efficiency (see Section \ref{ssec:efficiency}), speaks somewhat to the relevance for industry; additionally, some publications present industrial applications directly.
\citet{chandrashekaran2016automated} optimize the hyperparameters of their model for a large vocabulary continuous speech recognition with respect to two relevant performance metrics as well as memory footprint and show the results on a custom evaluation dataset provided by \textit{LGE Electronics}.
Another example can be found in \citet{quadrana2022multi}, which jointly optimizes for a number of relevant performance metrics in the context of behavioral song embeddings.
Results on two datasets are examined, one of which is a large-scale proprietary dataset of anonymized streaming listening sequences and playlists from \textit{Apple}.

\section{Discussion and open challenges}
\label{sec:discussion}
This paper has presented an overview of the concepts, methods and applications of MOHPO - as well as related concepts - and provides a guide to the ML practitioner delving into this particular topic.
It is evident there is merit to formulating ML problems in a multi-objective manner, as many application examples support.
Single-objective ML tasks, tuned to a pure prediction performance metric, are no longer in keeping with the state-of-the-art for many ML applications, as models have to meet certain standards with respect to secondary goals.
While this presents new challenges to the ML expert with regard to optimization and algorithm selection, proper methods can provide the user with a suite of Pareto optimal trade-offs to choose a suitable model.
It further enables practitioners to select meaningful and precise metrics as they no longer have to aggregate multiple metrics into a single one as is the case with e.g., AUC or F-score.
When utilizing MOO in this context such ``aggregate measures'' should be used with care; in general objectives should correspond to and mimic one real-world objective as closely as possible.
As the topic of MOHPO and multi-objective pipeline creation and model selection is not fully established, the available software (not for MOO, but specifically MOHPO) is limited, but good implementations exist for several standard methods.\\
\\
We conclude this work with a look at open challenges and directions for future work.
Integrating user preferences in a meaningful way either \emph{a priori} or during the optimization process (see Section~\ref{sssec:user_preferences}) remains a challenge that could help efficiency and transparency of MOHPO~\citep{pfisterer2019multiobjective}.
Especially obtaining (noisy) labels or user preferences in the process of MOHPO and utilizing them opens up a number of promising avenues for future research.
In some cases underlying user preferences may not be quantifiable and cannot be properly expressed through the metrics presented in Section~\ref{sec:applications}.
Future research in the direction of e.g. preference learning~\citep{gonzalez2017preferential, lin2022preference} or methods to integrate similar information into MOHPO is a topic worthy of additional exploration.
Another challenge is MOHPO beyond supervised learning: We have focused in this work on supervised learning, but unsupervised methods like anomaly detection or clustering also depend heavily on hyperparameters:
As in single-objective HPO~\citep{bischl2021hyperparameter}, the difficulty of performance evaluation and lack of standardized metrics complicate the application of the presented methods.
Aside from ``typical'' performance measures, other objectives like e.g., efficiency can still be concretely evaluated and may be included in (MO)HPO of unsupervised methods.
The fact, that essentially multiple black-box functions with individual characteristics (cost, noise etc.) and possible dependencies are being evaluated also presents a lot of opportunities.
If the cost of evaluation is drastically different between objective functions, this can be exploited to some extent: One option is to utilize hybrid methodologies combining e.g. model-based and evolutionary optimization techniques; hybrid algorithms in general are an open challenge for MOHPO applications.
Decoupled evaluations of objectives (with different costs) have been introduced for multi-objective Bayesian optimization~\citep{garrido2020parallel}, but application to MOHPO remains an open topic.
Another unsolved problem in this direction is to model the dependencies of various black-boxes.
Intuitively, if expert knowledge suggests that there is a correlation between the black-boxes, we could infer the value of one black box by knowing values of another black-box.
If exploited properly, this could allow forgoing evaluation of that black-box, especially if it will incur a significant loss of computational (or other) resources.
We motivated the general MOHPO problem to include constraints (see Section~\ref{mohpo_problem}), but in practice MOHPO is rarely combined with constraints and the topic of treating a desirable model quality as an objective vs. a constraint is hardly ever discussed.
Considering that these constraint - similar to the objective functions - may have different characteristics and be black-box, this presents a number of directions for future work.
Constraints and objectives are also sometimes hidden in MOHPO, which along with trustworthiness is one of the main reasons, why interactive methods are an interesting topic for future research.\\
\\
In general, a lot of open challenges should however be set in perspective to single-objective HPO: MOHPO has only recently been of increased interest and is therefore lacking behind in a number of respects.
This means, that some ideas for future work are simply catching up to the state-of-the-art in single-objective HPO.
A good example for this are certain algorithms, that only recently have been implemented for MOHPO (e.g., several multi-fidelity methods), but have been used productively in single-objective HPO for a number of years.
Simply looking at the unique challenges that HPO faces compared to other optimization problems, like high dimensional, mixed and hierarchical search spaces, showcases this further.
These issues have been discussed for single-objective HPO. Examples include special kernel functions or various surrogate models to tackle mixed-hierarchical hyperparameters~\citep{levesque2017conditional, daxberger2020mixedvariable, hutter2011sequential}
Additionally, a lot of the open challenges for single-objective HPO and single-objective AutoML like e.g., early stopping of optimization or noisy evaluations, need to also be considered for the multi-objective case.

Another point for future work we would like to emphasize is the lack of proper and extensive benchmarks for the field of MOHPO, which could shed some further light on strengths and weaknesses of different methods on a variety of MOHPO tasks.
Along the same lines, we encourage researchers to be transparent and exhaustive in the set-up and presentation of their experiments and benchmarks when it comes to introducing new methods; several works omit interesting baselines and do not detail the exact configuration of utilized algorithms among others.
This is further complicated by the fact that proper evaluation of MOHPO methods is still very much an open topic for future research.
Finally, the discrepancy in widespread vs. sparse use of MOHPO depending on the application at hand may also appear striking (see e.g., Section~\ref{ssec:mohpo_in_industry}). 
While ROC analysis and (multi-objective) feature selection are established and well researched areas; the body of literature for MOHPO including efficiency objectives has grown rapidly only in the past few years with the ascent of deep learning and HW-NAS.
With the recent trends to integrate FAT-ML related standards into the ML process, MOHPO with applications to interpretability and fairness is currently becoming increasingly relevant, but few works have been published that deal with these objectives as the community still struggles to establish metrics to quantify these important properties of ML models.
In regards to these objectives, it should be noted that MOHPO - compared to single-objective HPO - already improves in terms of transparency, simply by not having to reduce to a single metric and the result being a collection of trade-offs and not only a single hyperparameter configuration.
While there is plenty of research to be done in exploration of appropriate metrics for topics like interpretability or fairness, this is not necessarily an open topic for methodological MOHPO research\footnote{Other topics we have omitted here present further desirable model attributes, but currently no appropriate ways to translate these attributes into metrics exist. An example would be the amount of labeled data needed, which is an important concept in semi-supervised learning.}.

\begin{acks}
 This work was supported by the Bavarian Ministry of Economic Affairs, Regional Development and Energy through the Center for Analytics – Data – Applications (ADA-Center) within the framework of BAYERN DIGITAL II (20-3410-2-9-8) as well as the German Federal Ministry of Education and Research (BMBF) under Grant No. 01IS18036A.
\end{acks}

\bibliographystyle{ACM-Reference-Format}
\bibliography{main_arxiv}

\newpage 

\appendix

\section{List of Applications of MOHPO with at least one Objective Related to Efficiency}
\label{efficiency_applications_table}

\newcommand{\PreserveBackslash}[1]{\let\temp=\\#1\let\\=\temp}
\newcolumntype{C}[1]{>{\PreserveBackslash\centering}m{#1}}
\newcolumntype{R}[1]{>{\PreserveBackslash\raggedleft}p{#1}}
\newcolumntype{L}[1]{>{\PreserveBackslash\raggedright}p{#1}}

{\footnotesize
\renewcommand{\arraystretch}{1.7}
\begin{longtable}{C{0.6cm}m{3.8cm}m{3cm}m{2.65cm}m{2cm}}
Ref. & Optimization Method & Objectives & Task & Domain \\
\hline
\citep{hernandez2016designing} & PESMO (decoupled evaluations) & energy consumption, \newline prediction error & image classification \newline (MNIST)  & computer vision \\
\citep{parsa2019pabo}  & Bayesian Optimization (GP surrogate) for each objective  & energy consumption, \newline classification error & image classification \newline (CIFAR-10, flower17) & computer vision\\
\citep{reagen2017case} & PESMO & energy consumption, \newline classification error                                                         & image classification \newline (MNIST)                                             & computer vision             \\
\citep{hsu2018monas} & scalarization + reinforcement \newline learning & energy consumption or \newline MACs, accuracy  & image classification \newline (CIFAR-10) & computer vision \\
\citep{lu2019nsga}  & surrogate-assisted NSGA-II & FLOPs, classification error  & image classification \newline (CIFAR-10) & computer vision \\
\citep{gulcu2021multi} & Multi-Objective simulated \newline annealing & FLOPs, accuracy & image classification \newline (CIFAR-10) & computer vision \\
\citep{lu2020nsganetv2} & surrogate assisted NSGA-II & \# MAdds, accuracy & image classification \newline (CIFAR-10/100, MNIST) & computer vision \\
\citep{lu2020nsganetv2}  & (scalarization +) surrogate \newline assisted NSGA-II   & \# MAdds, accuracy  & image classification \newline (6 datasets) & computer vision \\
\citep{lu2020nsganetv2} & surrogate assisted NSGA-II & \# MAdds, \# parameters, \newline CPU latency, GPU latency, accuracy & image classification \newline (CIFAR-10/100, MNIST) & computer vision \\
\citep{shah2016pareto}                 & EIPV - Bayesian Optimization (multi-output GP with correlated objectives) EI in HV & memory consumption and training time (combined), accuracy  & classification \newline (boston housing) & - \\
\citep{chin2021joslim} & MO-BO (Gaussian processes) upper confidence bound and random scalarization & memory consumption or FLOPS, cross entropy loss & image classification \newline (CIFAR-10/100, \newline ImageNet) & computer vision \\
\citep{chandrashekaran2016automated} & scalarization + random search or BO (GP surrogate) or TPE or EA                                                         & peak memory consumpt., word error rate,  real time factor                                      & large vocab. continuous speech recognition                           & natural language processing \\
\citep{calisto2020adaen}   & EA based on MOEA/D  & \# trainable parameters, \newline segmentation accuracy & image segmentation  & medical \\
\citep{baldeon2020adaresu} & EA based on MOEA/D & \# trainable parameters, \newline segmentation accuracy  & image segmentation   & medical     \\
\citep{calisto2021emonas} & SaMEA: random forest surrogate assisted EA based on MOEA/D & \# parameters, expected segmentation error & image segmentation & medical \\
\citep{loni2019neuropower}           & SPEA2                                                                                                                       & \# parameters, accuracy                                                    & image classification \newline (CIFAR-10/100, MNIST)                        & computer vision             \\
\citep{liang2019evolutionary}        & CoDeepNEAT (MOEA) & \# parameters, accuracy & binary text classification (Wikidetox) & natural language processing \\
\citep{liang2019evolutionary}        & CoDeepNEAT (MOEA) & \# parameters, AUROC  & image classification  & medical   \\
\citep{deighan2021genetic}           & scalarization + EA variants                                                                                                 & network size, accuracy                                                            & gravitational wave classification                                        & physics                     \\
\citep{shinozaki2020automated}       & MO CMA-ES & network size, \newline word error rate & speech recognition  & natural language processing \\
\citep{juang2014structure}           & MO-RACACO \newline (Ant colony optimization)                                                                                         & network size (\# rule nodes), RMSE or SAE                                                 & time sequence prediction and nonlinear control problems                  & control design              \\
\citep{suttorp2006multobj_svm} & NSGA-II & \# support vectors, FPR, \newline FNR & pattern recognition & autonomous driving  \\
\citep{abdolshah2019preferences}     & Bayesian Optimization (GP surrogate) with (preference based) EHI as acquisition function & prediction time, \newline prediction error & image classification \newline (MNIST) & computer vision \\
\citep{hernandez2016predictive}      & PESMO                                                                                                                       & prediction time, \newline prediction error                                                                & image classification \newline (MNIST)                                             & computer vision             \\
\citep{koch2015efficient} & SMS-EGO variant with HV-based expected improvement & training time, \newline prediction accuracy & binary classification (Sonar from UCI repo.)  &  -  \\
\citep{tanaka2016automated} & MO CMA-ES & training time, \newline word error rate & speech recognition & natural language processing \\
\citep{qin2017evolution} & MO CMA-ES & validation time, translation performance (BLEU) & machine translation & natural language processing \\
\citep{rajagopal2020pso_efficiency} & Particle Swarm optimization & inference latency, accuracy & image/scene \newline classification & unmanned aerial vehicles \\
\citep{tan2019mnasnet} & weighted product method + \newline reinforcement learning & inference latency, accuracy   & image classification \newline (ImageNet) & computer vision \\
\citep{tan2019mnasnet}  & weighted product method + \newline reinforcement learning & inference latency, mAP & object detection \newline (COCO) & computer vision \\
\citep{belakaria2019maxvalue} & MESMO & prediction time, accuracy & image classification \newline (MNIST) & computer vision \\
\citep{kim2017nemo}                  & NSGA-II                                                                                                                     & prediction speed, accuracy                                                        & image classification \newline (MNIST, CIFAR-10)                                   & computer vision             \\
\citep{schmucker2021sh}              & Asynchronous successive halving                                                                                             & latency, accuracy                                                                 & image classification \newline (NAS-Bench-201)                                     & computer vision             \\
\citep{schmucker2021sh} & Asynchronous successive halving  & prediction time, perplexity, word error rate  & language modeling  & natural language processing \\
\citep{dong2018ppp-net} & Bayesian Optimization \newline (Recurrent Neural Network) & \# parameters, FLOPs, \newline prediction time, error rate & image classification \newline (CIFAR-10) & computer vision            
\end{longtable}}

\end{document}